\documentclass[DTMColor]{ROB-New}

\usepackage{siunitx}
\usepackage{acro}
\usepackage{dsfont}
\usepackage{amsfonts,amsmath,amssymb}
\usepackage{algorithm2e}
\usepackage{subcaption}
\usepackage{orcidlink}

\SetKwComment{Comment}{\# }{ }
\RestyleAlgo{ruled}

\usepackage{pifont}
\newcommand{\cmark}{\ding{51}}%
\newcommand{\xmark}{\ding{55}}%

\acsetup{list/template=description, case-sensitive=false, use-id-as-short=true}
\DeclareAcronym{cnn}{
  short = CNN ,
  long  = Convolution Neural Network ,
  pdfcomment  = Convolution Neural Network,
}

\DeclareAcronym{rpn}{
  short = RPN ,
  long  = Regional Proposal Network ,
  pdfcomment  = Regional Proposal Network,
}

\DeclareAcronym{3D}{
  short = 3D ,
  long  = three-dimension ,
  pdfcomment  = 3-dimension,
}

\DeclareAcronym{lidar}{
  short = LiDAR ,
  long  = Light Detection and Ranging ,
  pdfcomment  = Light Detection and Ranging,
}

\DeclareAcronym{yolo}{
  short = YOLO ,
  long  = You Only Look Once ,
  pdfcomment  = You Only Look Once,
}

\DeclareAcronym{ram}{
  short = RAM ,
  long  = Random-Access Memory ,
  pdfcomment  = Random-Access Memory,
}

\DeclareAcronym{CPU}{
  short = CPU ,
  long  = Computer Processing Unit ,
  pdfcomment  = Computer Processing Unit,
}

\DeclareAcronym{GPU}{
  short = GPU ,
  long  = Graphical Processing Unit ,
  pdfcomment  = Graphical Processing Unit,
}

\DeclareAcronym{fpga}{
  short = FPGA ,
  long  = Field Programmable Gate Array ,
  pdfcomment  = Fiel Programmable Gate Array,
}

\DeclareAcronym{hov}{
  short = HFOV ,
  long  = horizontal field of view ,
  pdfcomment  = Horizontal Field of View,
}

\DeclareAcronym{ann}{
  short = ANN ,
  long  = artificial neural network ,
  pdfcomment  = Artificial Neural Network,
}

\DeclareAcronym{som}{
  short = SoM ,
  long  = system on module ,
  pdfcomment  = System on Module,
}

\DeclareAcronym{dof}{
  short = DoF ,
  long  = degrees of freedom ,
  pdfcomment  = Degrees of Freedom,
}

\DeclareAcronym{svm}{
  short = SVM ,
  long  = support vector machine ,
  pdfcomment  = Support Vector Machine,
}

\begin{document}

\authormark{Magalhães, Sandro \textit{et al.}}
\copytext{Published in Robotica by Cambridge University Press}

\articletype{RESEARCH ARTICLE}

\jnlPage{1}{20}
\jyear{2024}
\jdoi{10.1017/S0263574724000936}

\title{MonoVisual3DFilter: 3D tomatoes' localisation with monocular cameras using histogram filters}

\author[1,2]{Sandro Costa Magalhães~\orcidlink{0000-0002-3095-197X}\hyperlink{corr}{*}}
\author[2]{Filipe Neves dos Santos~\orcidlink{0000-0002-8486-6113}}
\author[1,2]{António Paulo Moreira~\orcidlink{0000-0001-8573-3147}}
\author[3,4]{Jorge Dias~\orcidlink{0000-0002-2725-8867}}

\address[1]{Faculty of Engineering, University of Porto, Porto, Portugal}
\address[2]{INESC TEC -- Instituto de Engenharia de Sistemas e Computadores, Tecnologia e Ciência, Porto, Portugal}
\address[3]{Institute of Systems and Robotics, Department of Electrical Engineering and Computers, University of Coimbra, Coimbra, Portugal}
\address[4]{Khalifa University of Science, Technology, and Research, Abu Dhabi, United Emirates of Arabia (EUA)}
\address{\hypertarget{corr}{*}Corresponding author. \email{sandro.a.magalhaes@inesctec.pt}}

\received{xx xxx xxx}
\revised{xx xxx xxx}
\accepted{xx xxx xxx}


\abstract{
Performing tasks in agriculture, such as fruit monitoring or harvesting, requires perceiving the objects' spatial position. RGB-D cameras are limited under open-field environments due to lightning interferences. So, in this study, we state to answer the research question: ``How can we use and control monocular sensors to perceive objects' position in the 3D task space?'' Towards this aim, we approached histogram filters (Bayesian discrete filters) to estimate the position of tomatoes in the tomato plant through the algorithm MonoVisual3DFilter. Two kernel filters were studied: the square kernel and the Gaussian kernel. The implemented algorithm was essayed in simulation, with and without Gaussian noise and random noise, and in a testbed at laboratory conditions. The algorithm reported a mean absolute error lower than \SI{10}{\milli\metre} in simulation and \SI{20}{\milli\metre} in the testbed at laboratory conditions with an assessing distance of about \SI{0.5}{\metre}. So, the results are viable for real environments and should be improved at closer distances.
}

\maketitle

\section{Introduction}
\label{sec: introduction}

Agriculture is a critical sector in the global economy. Farmers and the agro-food industry have been adapting to meet the demands of the worldwide population, which is increasing fast \cite{Kitzes2007}. Several studies support that the population should keep increasing fast and reach about nine billion people by the year 2050 \cite{Perry2015,FAO2023}. Besides the increasing food demands to fulfil the global population \cite{FAO2023}, the area dedicated to agriculture can only increase marginally, requiring more optimised and precise strategies to improve production and cultivation ratios. These factors, associated with the labour shortage for agricultural tasks \cite{Leshcheva2018,Rica2015}, ally technology to farming and the agro-food industry.

Several scientific studies in the literature have been proving that robots can support farmers in agricultural tasks \cite{Schmitz2015,euRobotics2014,McBratney2005} and overcome the labour shortage. Mobile and intelligent robots can successfully perform tasks such as monitoring and harvesting. However, these robots require dedicated sensors to perceive fruits and other objects and estimate their localisation to the tools.

Recent literature reviews proved that most works for perceiving fruits use RGB-D sensors \cite{Magalhaes2022,Kumar2022,Nekoo2023,Colucci2023,Shen2017}. 
For instance, Sa \textit{et al.} \cite{Sa2017} performed a complete digitalisation of the scene to gather a digital twin of it and easily perceive the fruits and their \ac{3d} position. A \ac{svm}, based on colour and geometry features, performed a classification of the fruits. In another work, Jun \textit{te al.} \cite{Jun2021} used a YOLO v3 to detect the fruits in the scene, and, through an RGB-D camera, digitalised the fruit. Using this information, the authors built the Tool Centre Point algorithm to compute the centre of the fruit and the target point to harvest it. The point about these strategies is that they are being performed under controlled conditions at the laboratory or in controlled greenhouses.
Therefore, most of these essays were performed under controlled lightening, ensuring the sensors' correct functioning \cite{Kumar2022}. RGB-D sensors tend to malfunction under open-field environments due to reflections or intense illumination \cite{Kumar2022,Ringdahl2019,GeneMola2020}. So, using auxiliary algorithms and alternative technology is important to overcome the lightning effects. 

The previous conceptualisation permits the establishment of a common problem in open-field contexts, which we aim to approach in this study: ``How can we use and control monocular sensors to perceive the objects' positioning in the \ac{3d} task space?''.

Concerning estimating the depth using monocular cameras, most of the more recent works focused in \acp{cnn} to infer this relative depth to the sensor  \cite{Haq2022,Ma2020,Ma2019,Mousavian2017,Ranftl2022,Birkl2023}. Mousavian \textit{et al.} \cite{Mousavian2017} used a \ac{cnn} to estimate the \ac{3D} pose of an object and deployed a MultiBin loss function to optimise the model. Ma \textit{et al.} \cite{Ma2019,Ma2020} deployed custom-made \acp{cnn} named MonoPointNet and PatchNet to generate \ac{3D} images from monocular images and detect objects. Recently, Ranft \textit{et al.} \cite{Ranftl2022} and Birkl \textit{et al.} \cite{Birkl2023} released a family of \acp{cnn} based on MiDaS networks that aimed to estimate the relative depth to the cameras. The networks were trained and essayed on the set of the different datasets of RGB-D data in the literature. Also, Haq \textit{et al.} \cite{Haq2022} designed a new \ac{rpn} with geometric constraints to detect \ac{3D} objects using monocular cameras. This architecture performed similarly to \cite{Ma2019,Ma2020}. Van and Do \cite{Van2022} used a chessboard background and a regression-based \ac{cnn} to estimate the \ac{3D} pose of irregular objects using cuboids. The dependency on a chessboard background to predict the objects poses constraints on the model's applicability for unstructured environments. 

In the literature, we can also find purposeful deep learning solutions to extract the pose of the objects directly. The table \ref{tab: literature review} reports some examples of algorithms in the state of the art. SilhoNet \cite{Billings2019}, Nerf-Pose \cite{Li2023}, MORE \cite{Parisotto2023}, GhostPose \cite{Chang2021}, and GDR-Net \cite{Wang2021} are some of these solutions. SilhoNet reports an overall translation error of about \SI{2.45}{\centi\metre} using a complex state-of-the-art dataset containing multiple objects. Similarly, Nerf-Pose, MORE, GhostPose, and GDR-Net report an overall estimation error lower than \SI{2}{\centi\metre}. Collet and Srinivasa \cite{Collet2010} developed the introspective multiview algorithm that could estimate the pose of objects with estimation errors between \SIrange{0.46}{1.45}{\centi\metre}. These solutions illustrate remarkable results in the literature for objects' pose estimation. However, all of them are model-dependent in successfully estimating the pose of an object. An exception to this factor is the Imitrob \cite{Sedlar2023} that analyses the objects' configuration and structure and learns to estimate the pose of the objects from multiple perspectives.

\begin{table}[!htb]

\centering
\caption{Comparision with literature review}
\label{tab: literature review}

\begin{tabular}{@{}lcccc@{}}
\toprule
\multicolumn{1}{c}{\textbf{Algorithm}} &
  \textbf{Deep Learning} &
  \textbf{\begin{tabular}[c]{@{}c@{}}Model \\ dependency\end{tabular}} &
  \textbf{< \SI{2}{\centi\metre}} &
  \textbf{Error (\si{\centi\metre})} \\ \midrule
SilhoNet \cite{Billings2019}           & \cmark & \cmark & \SI{97.5}{\percent}  & --           \\
Nerf-Pose \cite{Li2023}          & \cmark & \cmark & --                   & \num{2.45}   \\
GDR-Net \cite{Wang2021}            & \cmark & \cmark & \SI{95.5}{\percent}  & --           \\
\cite{Collet2010}                    & \xmark & \cmark & --                   & \num{0.46} --  \num{2.45} \\
MORE \cite{Parisotto2023}  & \cmark & \cmark & \SI{93.94}{\percent} & --           \\
GhostPose \cite{Chang2021} & \cmark & \cmark & \SI{93.9}{\percent}  & --           \\
Imitrob \cite{Sedlar2023}  & \cmark & \xmark &        --            & \num{6.5}    \\ \bottomrule
\end{tabular}
\end{table}

Despite being largely explored in the literature, deep learning-based solutions are data exhaustive, requiring much and varied data with their features well identified and represented. Acquiring data to train deep learning models is expensive and difficult. For natural environments, that requires going to the natural scenes and mapping the objects for the robot's context.

Other works used auxiliary sensors to create \ac{3D} scenes or depth estimation, such as \acp{lidar} \cite{Iyer2018}. However, high resolution and quality \acp{lidar} are expensive, and increase the effort over robotic manipulators to manoeuvre and pick objects and also constrained by their operation in open-field robotics.    

There are still some researchers that opted for using monocular cameras and controlling the path to the objects using visual servoing strategies \cite{Qu2017,Kuecuek2004}. Shen \textit{et al.} \cite{Shen2017} applied visual servoing using RGB-D cameras. A distinctive work is presented by Xu \textit{et al.} \cite{Xu2020} which used the brightness of the environment and the movement of the camera to estimate the depth and reconstruct the structure of the colon during endoscopies.

Probabilistic algorithms are also capable of estimating the objects' poses accurately. Algorithms such as Bayesian Histogram Filters were explored in the literature to identify and localise objects in the scenes using different kinds of sensors. Bayesian Histogram Filters are commonly used in robotics for localisation navigation purposes \cite{Thrun2005,Boyko2021,Moscowsky2021}, but their potentialities enable him for other aims such as identifying and localising other objects. Sarmento \textit{et al.} \cite{Sarmento2022} applied region histogram filters to identify people, animals and other obstacles in the ultrawideband scene to avoid collisions and follow people. Also, Engin and Isler \cite{Engin2021} used this algorithm to localise random objects. Márton \textit{et al.} \cite{Marton2017} used a histogram filter to complement the information provided by a state-of-the-art position estimator and accurately estimate the object's orientation.

The advantage associated with histogram filters is that they are only constrained by the detection capabilities of the algorithms behind the sensors. Therefore, using a well-defined model to perceive the fruits through monocular images, we can translate their 2D relation to the 3D and accurately estimate the objects' positions. Besides, the histogram filters propose a working philosophy similar to triangulation, which has their accuracy and functionality well proved in geospatial and cartography disciplines.

\subsection{System and requirements}

Greenhouses are a target-designed scene to optimise the production of fruits in environment-controlled conditions. They also have the advantage that the crops can be modelled to better fit the farm and objectives constraints. So, besides the complexity of agricultural scenes, the modelisation of the environment can be simplified in greenhouses, if robust perception algorithms are used. Besides, the greenhouse context actually has the purpose and the need to adapt to human and robotic systems \cite{vanHerck2020,Bac2014}, becoming more effective and ergonomic for different operations.

\begin{figure}[!htb]
    \centering
    \includegraphics[width=0.3\textwidth]{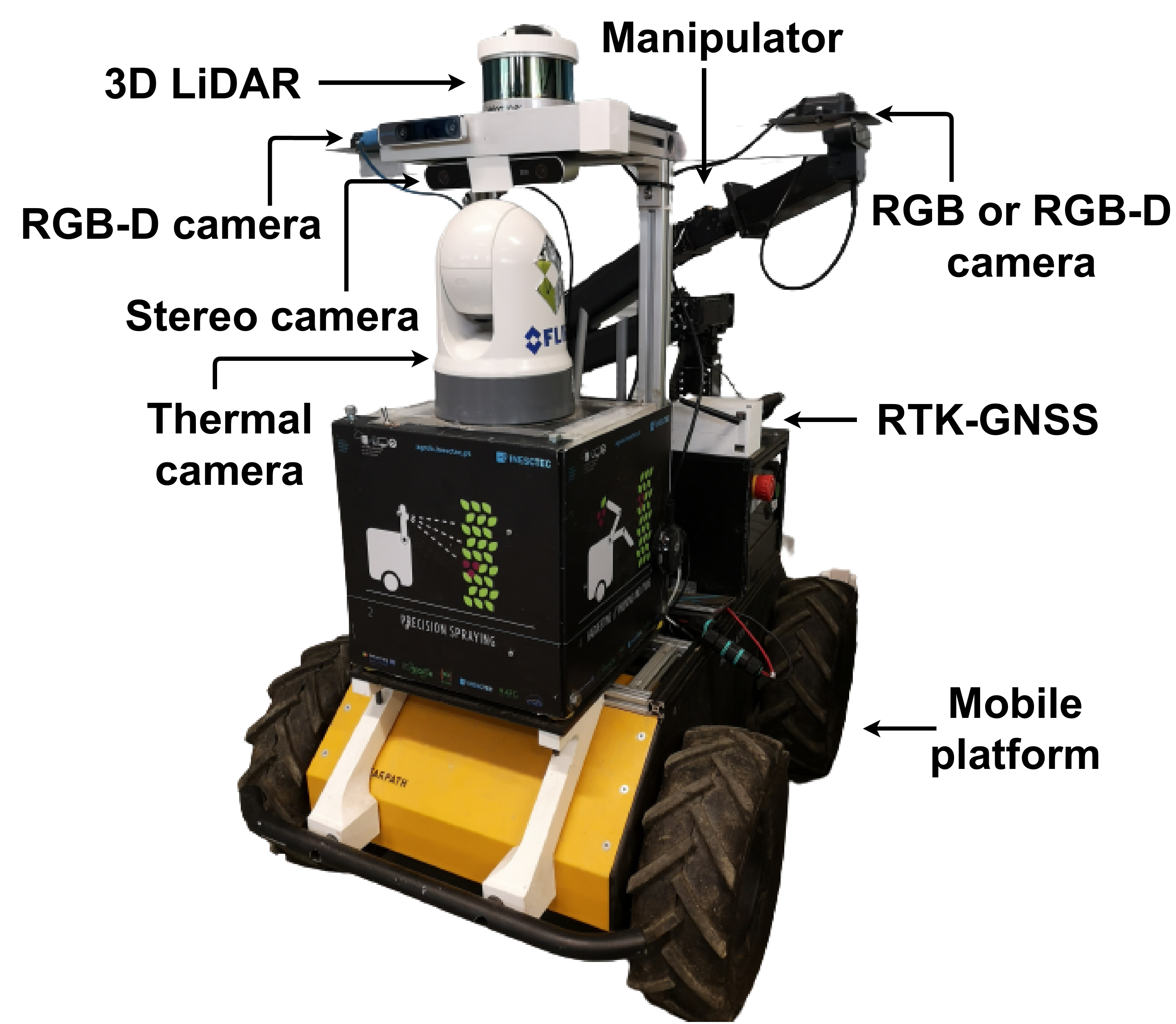}
    \caption{Robot AgRob v16 from INESC TEC to operate under open-field and controlled agricultural environments.}
    \label{fig:agrob}
\end{figure}

Actual robotised greenhouses are usually operated by robots on trails \cite{Arad2020}. However, these environments require the development of robot-specific environments and do not fit in the commonly operated greenhouses. Therefore, robots operating in the most common greenhouses should be composed of wheeled mobile robots. AgRob v16 from INESC TEC (Fig. \ref{fig:agrob}) is a wheeled robot with the Clearpath Husky platform\footnote{See Clearpath Robotics 2023. Husky -- Unmanned Ground Vehicle. Online \url{https://clearpathrobotics.com/husky-unmanned-ground-vehicle-robot/} [Last accessed on May 12th, 2023]} and all-terrain wheels designed for operating under open-field and controlled agricultural environments. This robot is currently being essayed in Douro steep slope vineyards and tomato greenhouses. It has a perception and controlling head that gathers data and information from the environment for navigation and mapping. Additionally, the Robotis Manipulator-H was assembled at the backside to perform tasks in the cultivars, such as monitoring, pruning, or harvesting.

Perception for manipulation is usually performed through eye-in-hand techniques. The Robotis Manipulator-H can handle either monocular or RGB-D cameras. RGB-D cameras can perceive the \ac{3D} pose of objects, but they are bigger and more difficult to handle. Besides, RGB-D cameras have difficulty perceiving and estimating the objects' depth in open-field environments due to natural lightning interferences \cite{Kumar2022,Ringdahl2019,GeneMola2020}. Therefore, using monocular cameras, the manipulator can perceive the position of the objects from multiple perspectives and estimate the objects' pose. Approaches based on machine learning or statistics are commonly explored in the literature \cite{Haq2022,Ma2020,Ma2019,Mousavian2017}, albeit statistical approaches are more analytical solutions and with more predictable results.

\vspace{24pt}

Therefore, as reviewed, Bayesian Histogram Filters are suitable for identifying the \ac{3D} position of objects and do not require the model's knowledge. Besides, the histogram filters are more predictable and explainable than deep learning-based solutions, which makes it also easier to readjust to new scenarios. So, for this work, we applied the histogram filter to identify the \ac{3D} position of tomatoes in a testbed using a monocular camera assembled in a robotic arm, in a solution called MonoVisual3DFilter. At the current stage, the arm used fixed multi-viewpoint to observe the tomatoes from multiple perspectives. 

The current work benefits, in our knowledge, for the first essay in using Bayesian Histogram Filters for estimating the \ac{3D} position of objects in the range of a robotic manipulator using monocular cameras. The implemented algorithm was evaluated for fruit detection under simulation and testbed conditions in the laboratory. Therefore, the current work contributes by:

\begin{itemize}
    \item Introducing and essaying the MonoVisual3DFilter;
    \item First, apply histogram filters to detect the centre position of objects;
    \item Apply the algorithm to real-world problems, such as fruit detection in the plants' canopy; and
    \item Study the effect and advantages of different kernel types.
\end{itemize}

The following sections are structured: material and methods, results, discussion and conclusion. In section \ref{sec: materials and methods}, we detail the conditions of the experiments and formalise the algorithm application. Section \ref{sec: results} introduces the different results for the different experiments, which are analysed and discussed, in section \ref{sec: discussion}, and concluded in section \ref{sec: conclusion}.

\section{Materials and Methods}
\label{sec: materials and methods}

\subsection{Real data and simulation}

The development and the experiments with the algorithm MonoVisual3DFilter were done under two environments: simulation and a testbed in the laboratory (near real-world conditions).

\begin{figure}[!hbt]
    \centering
    \begin{subfigure}[b]{0.4\textwidth}
         \centering
         \includegraphics[width=\textwidth]{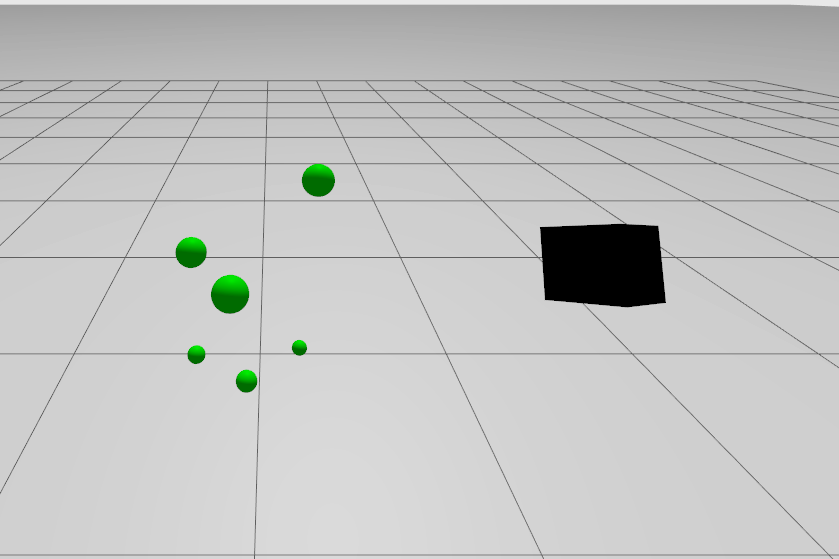}
         \caption{}
    \end{subfigure}
    \begin{subfigure}[b]{0.4\textwidth}
         \centering
         \includegraphics[width=0.9\textwidth]{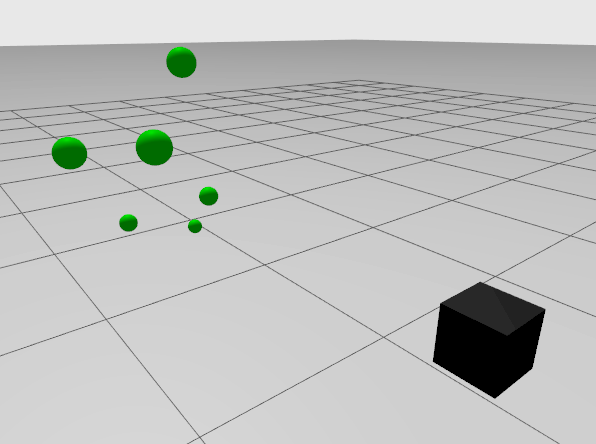}
         \caption{}
    \end{subfigure}
    \caption{Simulated environment to validate the histogram filter effectiveness. Green spheres are the objects being detected, representing the tomatoes, and the black box is the bounding box camera looking at the spheres.}
    \label{fig:simulation}
\end{figure}

A simplistic simulation environment was designed using the Ignition Gazebo Simulator\footnote{See Open Robotics, ``Gazebo,'' accessed on May 12th, 2023. [Online]. Available: \url{https://gazebosim.org/}}. The scene comprises six spheres to assess the algorithm's validity and test during implementation (Fig. \ref{fig:simulation}). The spheres have sizes of \SI{5}{\centi\metre} and \SI{10}{\centi\metre}. A bounding box camera was added to the scene to perceive the objects and support the position estimation algorithm. During the execution of the MonoVisual3DFilter (a histogram filter), the bounding box camera is moved to fixed viewpoints to enable and validate the estimator.  The bounding box camera detects the objects through object detection algorithms using bounding boxes, detecting only their visible regions.

\begin{figure}[!htb]
    \centering\hfill
    \begin{subfigure}[b]{0.27\textwidth}
         \centering
         \includegraphics[width=\textwidth]{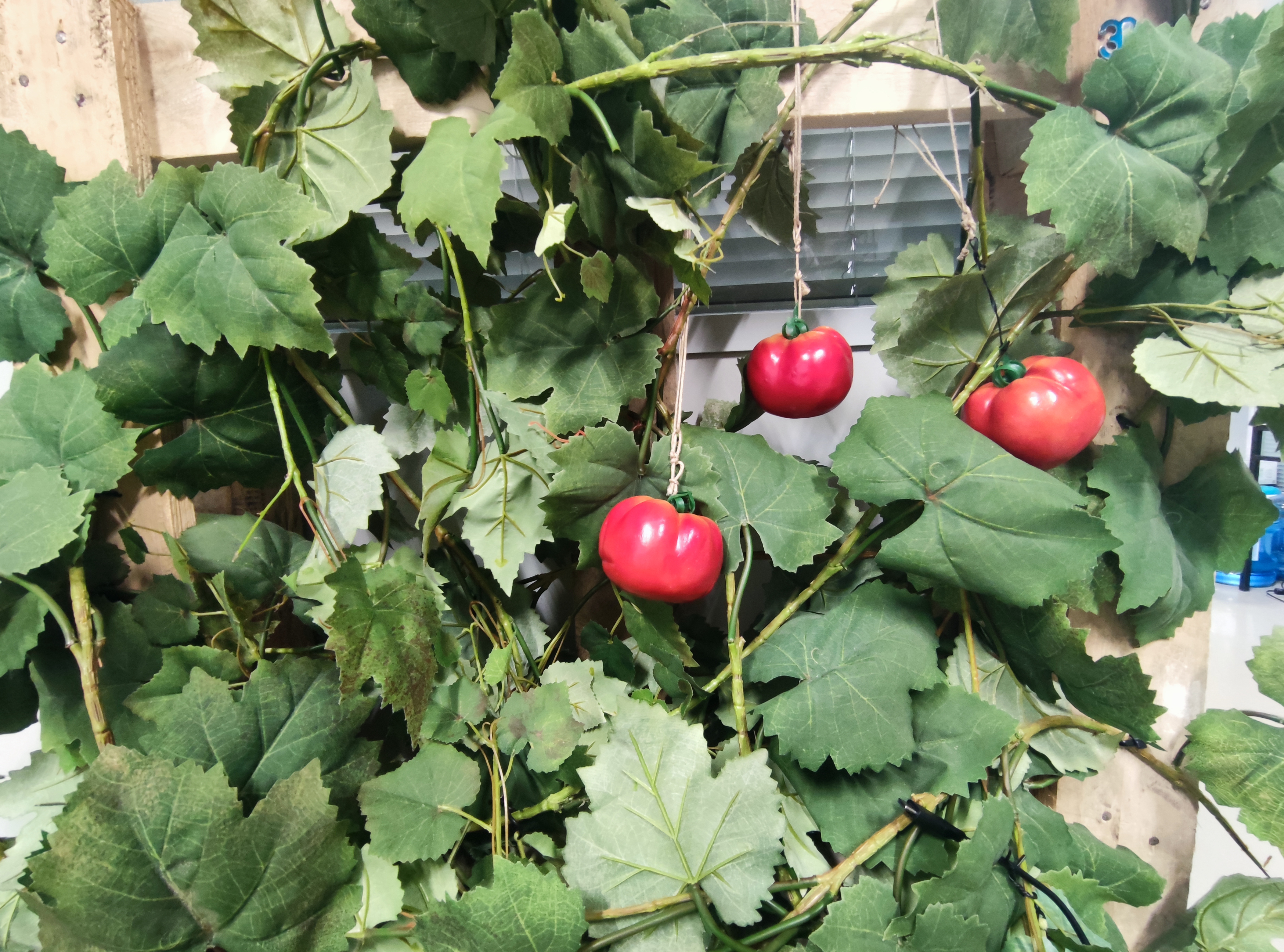}
         \caption{Testbed with the tomatoes}
         \label{fig:testbed}
    \end{subfigure}\hfill
    \begin{subfigure}[b]{0.20\textwidth}
         \centering
         \includegraphics[width=\textwidth]{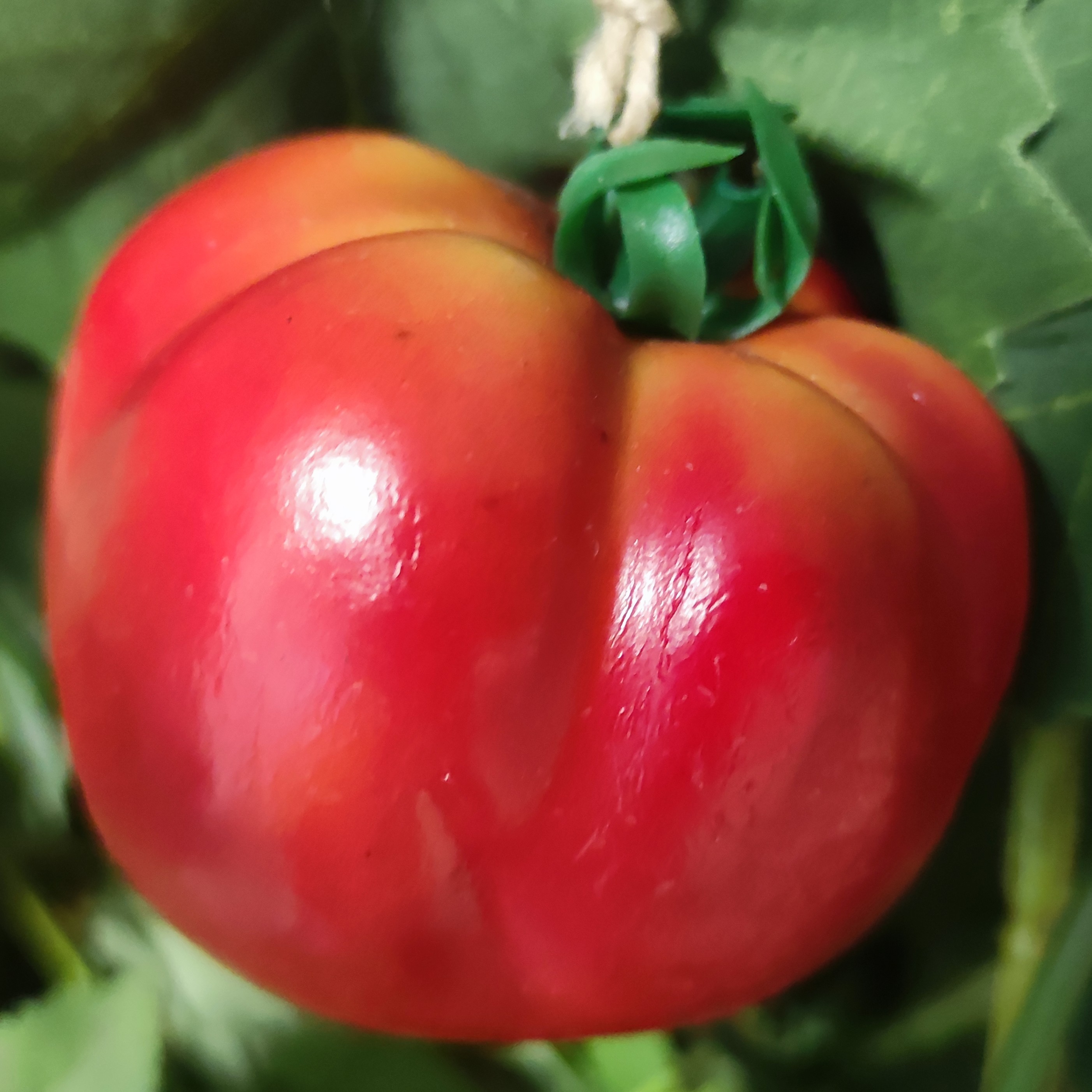}
         \caption{Plastic tomato}
         \label{fig:tomato}
    \end{subfigure}\hfill
    \begin{subfigure}[b]{0.20\textwidth}
         \centering
         \includegraphics[width=\textwidth]{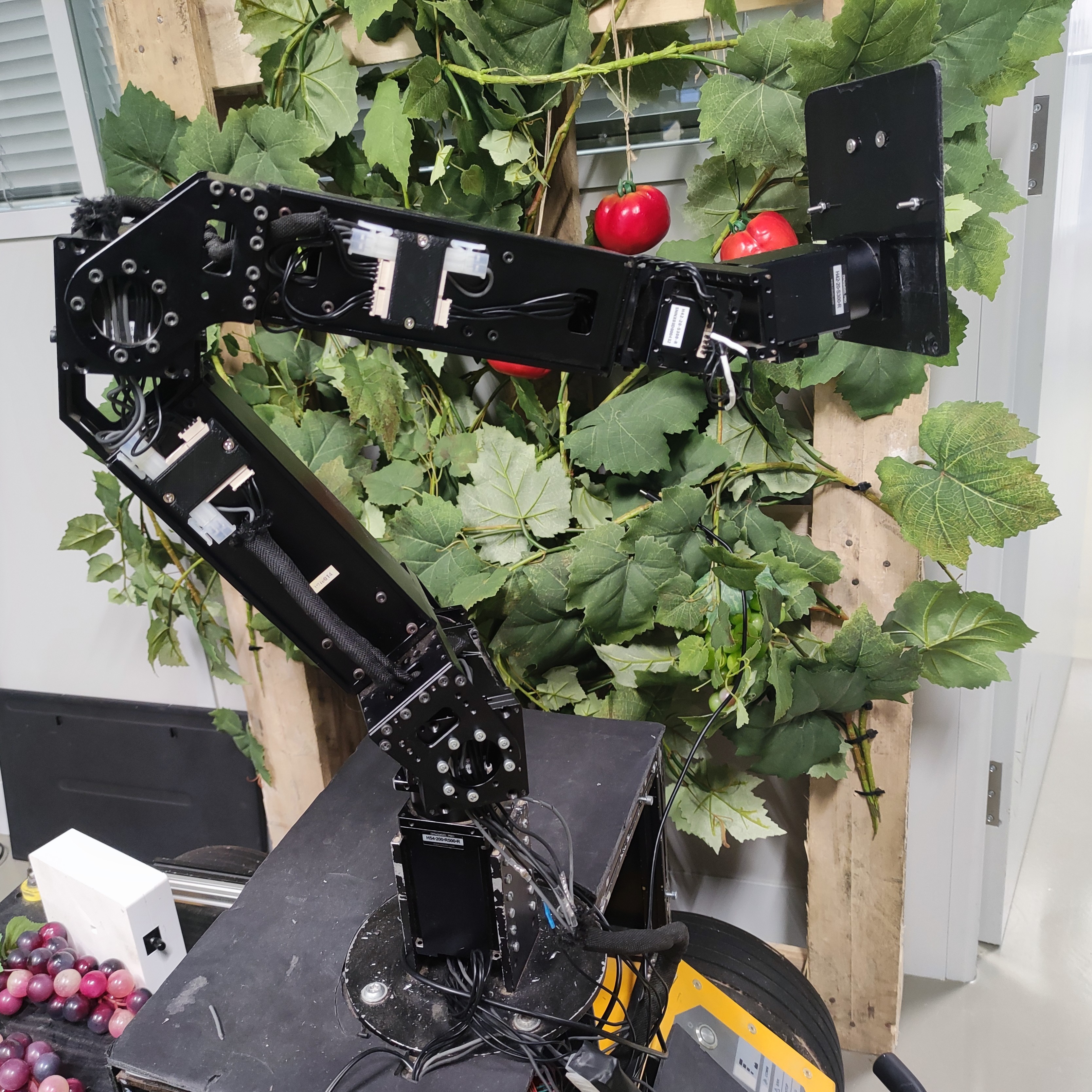}
         \caption{Manipulator-H}
         \label{fig:manipulator}
    \end{subfigure}\hfill
    \begin{subfigure}[b]{0.20\textwidth}
         \centering
         \includegraphics[width=\textwidth]{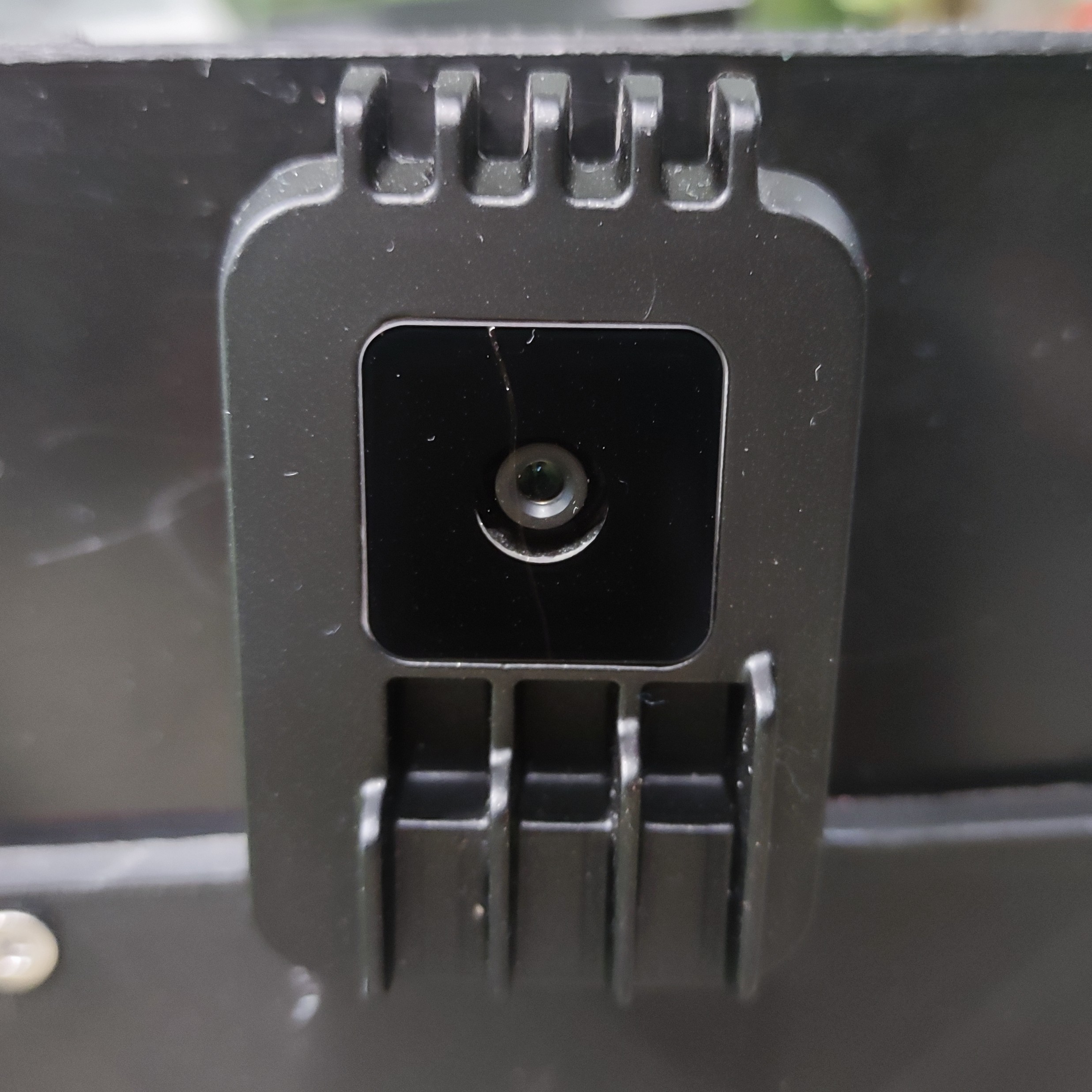}
         \caption{OAK-1 Camera}
         \label{fig:oak}
    \end{subfigure}\hfill\hfill
    \caption{Simulated testbed in the laboratory to essay the histogram filter algorithm}
    \label{fig:testbed-lab}
\end{figure}

To essay the algorithm in the laboratory, near real-world greenhouse conditions, we designed a testbed with realistic leaves and plastic realistic tomatoes (Fig. \ref{fig:testbed} and \ref{fig:tomato}). For perceiving the tomatoes, we assembled an OAK-1 camera\footnote{See Luxonis Holding Corporation, ``OAK-1,'' accessed on September 26th, 2023. [Online]. Available: \url{https://docs.luxonis.com/projects/hardware/en/latest/pages/BW1093/}} (Fig. \ref{fig:oak}), as a bounding box camera, to the 6 \ac{dof} Robotis Manipulator-H (Fig. \ref{fig:manipulator}). The OAK-1 camera computes an object detection model algorithm trained to perceive tomatoes. We used the \ac{yolo} v8 Tiny trained on the tomato dataset \cite{Magalhaes2021,Magalhaes2020} and some samples of the plastic tomato from multiple perspectives. However, any object detection or instance segmentation algorithm can be used to perceive the objects of interest in the scene, since they can effectively lead with the different environment perturbances, such as lighting variations or are robust to occlusion. The manipulator also moved to fixed viewpoints that ensured the tomatoes' visibility. The manipulator was assembled on the mobile platform AgRob v16 from INESC TEC (Fig. \ref{fig:agrob}), but \ac{3D} position of tomatoes was computed to the manipulator's base frame.

The OAK-1 is a fully integrated system for bounding box cameras. This camera was designed by the Luxonis Holding Corporation and has a single \SI{12}{MP} RGB camera module. To allow the on-system object detection, the camera module is connected to the OAK-SoM\footnote{See Luxonis Holding Corporation, ``OAK-SoM,'' accessed on September 26th, 2023. [Online]. Available: \url{https://docs.luxonis.com/projects/hardware/en/latest/pages/BW1099/}}. The full camera connects to other devices through USB-C communication. The OAK-SoM is a \ac{som} designed to integrate into top-level and low-power systems and has the capability to process \acp{ann}. This camera module was integrated into the AgRob v16 robot (Fig. \ref{fig:agrob}).

This robot is based on the Clearpath Husky\footnote{See Clearpath Robotics Inc., ``Husky - Unmanned Ground Vehicle,'' accessed on September 26th, 2023. [Online]. Available: \url{https://clearpathrobotics.com/husky-unmanned-ground-vehicle-robot/}} mobile platform and was designed to operate in harsh agricultural environments, such as the Douro's steep slope vineyards. At the front of the mobile platform, there is a controlling head, which is the unit with the computer and all the required devices to control the robot and establish communications. At the backside, the robot has the 6 \ac{dof} anthropomorphic manipulator Robotis Manipulator-H\footnote{Robotis, ``Robotis e-Manual -- Manipulator-H,'' accessed on September 26th, 2023. [Online]. Available: \url{https://emanual.robotis.com/docs/en/platform/manipulator_h/introduction/}}.

\subsection{Histogram Filter}

Histogram filters have been widely used in literature for self-localisation and navigation in mobile robots \cite{Boyko2021,Moscowsky2021}. However, for the current study, we intend to apply histogram filters to localise the \ac{3D} position of tomatoes concerning the manipulator's base frame, in a solution called MonoVisual3DFilter.

\begin{figure}[!htp]
    \centering
    \includegraphics[width=5cm]{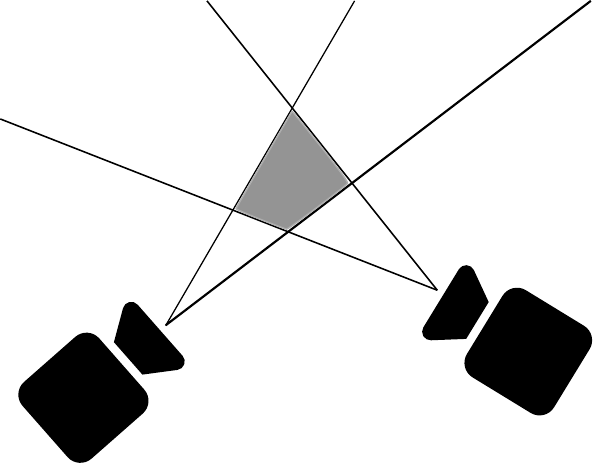}
    \caption{Intersection between multiple viewpoints in 2D plane}
    \label{fig:intersection_spaces}
\end{figure}

The histogram filter is a computationally intensive algorithm that can estimate the relative position of objects. The filter computes probabilities for multiple points in a discretised space. After, it intersects the chances of various views to estimate the localisation and the occupied area of the regions of interest (Fig. \ref{fig:intersection_spaces}).

Histogram filters can be set as an application of a discrete Bayes filter to the continuous state space. For this study, we applied the histogram filter as stated by Thrun~\cite{Thrun2005}.

The histogram filter decomposes the continuous state space in a finite number of regions. Equation \ref{eq: discrete state space} describes a discretised state space. $X_t$ is the random variable describing the state of the objects being detected at the time $t$. $\text{dom}(X_t)$ denotes the state space{, which is all the possible values that $X_t$ might assume}. The most straightforward discretisation of a continuous state space is through a multidimensional grid, where $x_{k,t}$ denotes each grid cell.

\begin{equation}
    \text{dom}(X_t) = x_{1,t} \cup x_{2,t} \cup \dots \cup x_{K,t}
    \label{eq: discrete state space}
\end{equation}

Only part of the state space must be discredited to limit computation efforts, mainly due to the manipulator's reachability. Therefore, as soon as the manipulator's camera detects an object of interest, in the first viewpoint, the space behind the camera is decomposed through a grid scheme. We considered the probable space around the object as twice the manipulator's reaching limits. Twice the manipulator's reachability offers enough margin to identify and validate some fruits in the limits of the manipulator's reachability. The \ac{3D} decomposition is centred in the camera in $y\mathcal{O}z$, i.e. in $(0, 0)$ in the camera's frame and further distanced by the manipulator's reachability radius in $\mathcal{O}x$. Once decomposed, the discrete state space remains static and only $\text{dom}(X_t)$ is updated.


The histogram filter demands moving the camera to multiple strategic viewpoints and updating the probabilities grid. So, at the space decomposition, an associated probabilities matrix is created with a probability to each cell initialised to one, i.e. $\text{dom}(X_0) = [1 \dots 1]$. This means that at the beginning, the object of interest is probable to be anywhere in the decomposed space. 

For estimating the position of the objects, $\text{dom}(X_t)$ is updated in each viewpoint. Each cell, $x_{i,t}$, from the decomposed state space is transformed from the manipulator's base frame to the image's frame. The probability of an object in a given cell, $x_{i,t}$, knowing the viewpoint, is given by \eqref{eq: create probability}. Finally, the updated probability of an object being in the cell $x_{i,t}$ is given by \eqref{eq: update cell probability}.

\begin{align}
    p(x_{i,t} | \text{viewpoint}_k) = & \dfrac{1}{N} \sum_j^N p(x_{i,t} | \text{bbox}_j, \text{viewpoint}_k) \label{eq: create probability}\\
    p(x_{i,t}) = & p(x_{i,t} | \text{viewpoint}_k) \cdot p(x_{i,t-1}) \label{eq: update cell probability}
\end{align}

In equation \ref{eq: create probability}, to get $p(x_{i,t} | \text{bbox}_j, \text{viewpoint}_k)$, a kernel function was designed. Two kernel functions were essayed to localise the objects in the state space: the square and Gaussian functions.
The square kernel, applied to each bounding box, states that if the transformed point is inside a bounding box of the image's frame, the probability is one; otherwise is zero \eqref{eq:square_kernel}. Using the square kernel function, we will have a binary mask stating that if a point is inside the bounding box, then we can have an object. Otherwise, we don't. 

\begin{equation}
    p(x_{i,t} | \text{bbox}_j, \text{viewpoint}_k) = \begin{cases}
        1 & \text{if inside bounding box} \\
        0 & \text{otherwise}
    \end{cases}
    \label{eq:square_kernel}
\end{equation}

As an alternative to the aggressive behaviour of the square function, we also essayed the bidimensional Gaussian function \eqref{eq:gaussian_kernel}. This function delivers a smooth effect for the borders of the bounding box and some cells $x_{i,t}$ outside the bounding boxes. Therefore, a Gaussian function should better tolerate irregular objects and noise. In the equation \ref{eq:gaussian_kernel}, $(x_0, y_0)$ is the centre of bounding box $j$ in the sensor's frame and $(x,y)$ is the position of each cell $x_{i,t}$ in the sensor's frame.  The coordinates in the image's frame are projected by a projection model stated in section \ref{sec: camera model}. The standard deviation values $(\sigma_x, \sigma_y)$ correspond to half of the size of the bounding box $j$. All the values were obtained experimentally and had reasonable results. If we use the Gaussian kernel (\ref{eq:gaussian_kernel}) to estimate the objects' position, equation \ref{eq: create probability} will correspond to a mixture of Gaussians, attending that the detection camera detects multiple objects. The mixture of Gaussians results in a function that smooths with increasing Gaussians in the mixture. To avoid this effect, a normalised version of the mixture of Gaussians is used to highlight the different detected objects \eqref{eq:normalise_gaussian_kernel}. Besides, the updated state space $\text{dim}(X_t)$ is also normalised at the end of each iteration of the histogram filter. 

\begin{align}
    p(x_{i,t} | \text{bbox}_j, \text{viewpoint}_k) = & \exp{\left(-\dfrac{(x-x_0)^2}{2\sigma_x^2}-\dfrac{(y-y_0)^2}{2\sigma_y^2}\right)}
    \label{eq:gaussian_kernel} \\
    p(x_{i,t} | \text{viewpoint}_k) = & \dfrac{p(x_{i,t} | \text{viewpoint}_k)}{\max(p(x_{i,t} | \text{viewpoint}_k))} \label{eq:normalise_gaussian_kernel}
\end{align}

The algorithm \ref{alg:histogram_filter} summarises the procedures for updating the cell's weights during the histogram filtering. 

\begin{algorithm}
\caption{Histogram filter -- updating weights} \label{alg:histogram_filter}
\KwData{$decomposition\_grid$, $probabilities\_matrix$, $bounding\_boxes$}
\KwResult{$probabilities\_matrix$}
\For {each viewpoint}{
    \For {$x_{i,t}, p(x_{i,t})$ \texttt{in} $decomposition\_grid$, $probabilities\_matrix$}{
        $cell\_camera \gets$ \text{transforms cell from the mainframe to camera's frame}\;
        $cell\_sensor \gets cell\_camera$ \text{in the sensor's frame}\;
        $(u,v) \gets cell\_sensor$ \text{in the image's frame}\;
        $p(x_{i,t} | \text{viewpoint}_k) \gets$ 0 \;
        \For{bbox \texttt{in} $bounding\_boxes$}{
            $p(x_{i,t} | \text{viewpoint}_k) \gets p(x_{i,t} | \text{viewpoint}_k) + \dfrac{1}{N} \times p(x_{i,t} | \text{bbox})$\;
        }
    }
    $p(x_{t} | \text{viewpoint}_k) \gets \text{normalise}(p(x_{t} | \text{viewpoint}_k))$\;
    $p(x_{t}) \gets p(x_{t}) \times p(x_{t} | \text{viewpoint}_k)$\;
    $p(x_{t}) \gets \text{normalise}(p(x_{t}))$\;
}
\end{algorithm}

\subsection{Camera Projection Model}
\label{sec: camera model}

While applying the histogram filter, we decomposed the state space in a finite state space grid. To effectively estimate the \ac{3D} position of the object, we moved the detection camera around the object and the decomposed state space to visualise the scene from multiple perspectives. 

\begin{figure}[!htb]
    \centering
    \includegraphics[width=5cm]{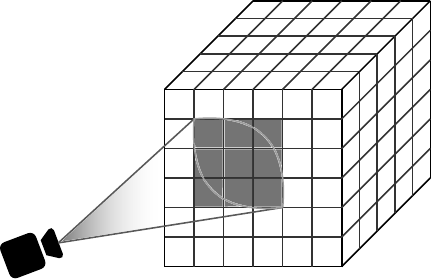}
    \caption{Intersection of the camera around in the decomposed space.}
    \label{fig:camera_space}
\end{figure}

Detecting objects using the detection camera requires an effective projection model to estimate the object's position in the \ac{3D} space, transforming between the \ac{3D} space coordinates and the image's frame. For simplification, we applied the Pinhole model to transform the \ac{3D} space coordinates to the image's frame.

\begin{figure}[!htb]
    \centering
    \includegraphics[width=4cm]{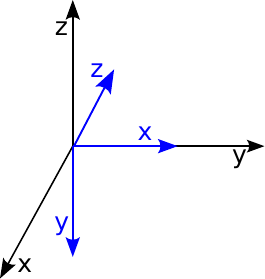}
    \caption{Conversion between the camera's and sensor's frames (blue -- sensor's frame; black -- camera's frame).}
    \label{fig:camera_sensor_frame}
\end{figure}

Acknowledging the points of the 3D space in the camera's frame, before we project them in the image's frame, we have to convert them to the sensor's frame. Both are placed in the same origin, but they have different orientations. The sensor uses the frame as illustrated in Fig. \ref{fig:camera_sensor_frame}. The illustrated rotation can be stated by Euler angles like Euler(YZX) $=$ (\SI{0}{\degree}, \SI{90}{\degree}, \SI{-90}{\degree}) which reflects in the quaternion $q=(-0.5, 0.5, -0.5, 0.5)$.

The transformation between the sensor's frame and the image's coordinates can be made by the intrinsics parameters matrix as stated in equation \ref{eq:intrinsics}. This matrix depends on the image's width ($w$) and height ($h$), as well as on the camera's focal length ($f$). The focal length depends on the camera's \ac{hov} and the image's width and can be calculated by equation \ref{eq:field_of_view}. 

\begin{equation}
    (u,v,1) = \begin{bmatrix}
        f & 0 & \frac{w}{2} \\
        0 & f & \frac{h}{2} \\
        0 & 0 & 1 \\
    \end{bmatrix} \cdot \begin{bmatrix}
        \frac{x}{z} \\ \frac{y}{z} \\ 1
    \end{bmatrix}
    \label{eq:intrinsics}
\end{equation}

\begin{equation}
    f = \dfrac{0.5 \times w}{\tan\left(0.5 \times HFOV \times \dfrac{\pi}{180}\right)}
    \label{eq:field_of_view}
\end{equation}

However, this model type is only valid behind ideal scenes, such as in the simulation. For the OAK-1 camera, an additional calibration step was required to estimate the intrinsic parameters. For calibration, we used the Kalibr software \cite{Maye2013}.

\subsection{Objects positioning}

At the end of the execution of the histogram filter for multiple viewpoints, the state space $\text{dom}(X_t)$ should have different clusters of points. As we know the number of objects in the scene through the number of detected objects by the detection camera, we can use the k-means algorithm to aggregate the points and compute the centre of each cluster. 

The k-means algorithm tries to cluster the different points of the discrete state space by minimising the geometric distance between points to the cluster's centre \eqref{eq: k-means}. In the equation \ref{eq: k-means}, we minimise the Euclidean distance between points to $\mu_j$, the centre point of each cluster in $C$.

\begin{equation}
    \sum_{i=0}^{n} \min_{\mu_j \in C}(||x_{i,t} - \mu_j||^2)
    \label{eq: k-means}
\end{equation}

After clustering by the k-means, the state space should have as many point clouds as the number of objects detected by the detection camera. The computation of the centre of the clouds to get the position of the detected objects can be done in two ways: 

\begin{enumerate}
    \item computation of the geometric centre of the cloud; or
    \item computation of the weighted centre of the cloud. 
\end{enumerate}

The geometric centre of each point cloud is the Euclidean centre $\mu_j$ minimised during the k-means algorithm for the equation \ref{eq: k-means}. Besides the geometric centre, the k-means also return the points, $x_{i,t}$, that belong to each cloud, $S_j$. Considering the state of each element of the state space $\text{dom}(X_t)$ at the end of the histogram filter, each element $x_{i,t}$ should have a weight attributed, $w_i$. So the weighted centre is the weighted average \eqref{eq: weighted centre} of the coordinates of $x_{i,t}$ that belong to the set $S_j$. 

\begin{equation}
    \mu_j = \begin{bmatrix}
        \dfrac{\sum_i^N w_i \cdot x_{{i,t}_1}}{N} &
        \dfrac{\sum_i^N w_i \cdot x_{{i,t}_2}}{N} &
        \dfrac{\sum_i^N w_i \cdot x_{{i,t}_3}}{N}
    \end{bmatrix}^T \qquad \forall x_{i,t} \in S_j
    \label{eq: weighted centre}
\end{equation}

\subsection{Experiments}

Three essays were performed in different environments to validate the effectiveness of the MonoVisual3DFilter. 

\begin{figure}[!htb]
    \hfill%
    \begin{subfigure}[b]{0.24\textwidth}
         \centering
         \includegraphics[width=\textwidth]{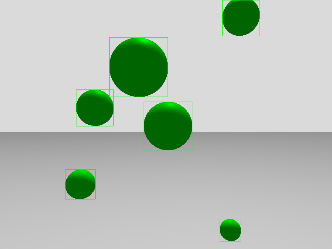}
         \caption{}
         \label{fig:sim_cam_p1}
    \end{subfigure}
    \hfill%
    \begin{subfigure}[b]{0.24\textwidth}
         \centering
         \includegraphics[width=\textwidth]{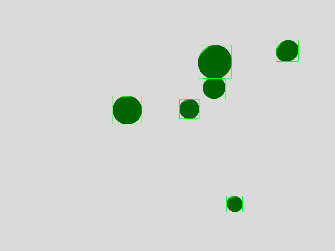}
         \caption{}
         \label{fig:sim_cam_p2}
    \end{subfigure}
    \hfill%
    \begin{subfigure}[b]{0.24\textwidth}
         \centering
         \includegraphics[width=\textwidth]{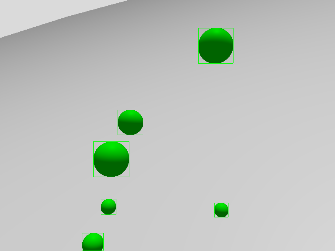}
         \caption{}
         \label{fig:sim_cam_p3}
    \end{subfigure}
    \hfill%
    \hfill%
    \caption{View of the spheres by the bounding box camera at each fixed viewpoint. The green square boxes around the spheres are the bounding boxes of the detected spheres by the bounding box camera.}
    \label{fig:sim-cam}
\end{figure}

Using the Gazebo simulator, we created a scene with multiple spheres to estimate their position in the scene (Fig. \ref{fig:simulation}). Once we used a bounding box camera without noise, this approach allowed validating the real performance of the filter without external artefacts or noise. Additionally, we performed an additional essay, introducing some Gaussian noise that randomly changes the position and size of the bounding box of the detected objects, as well as whether the object is successfully detected. Because we do not assemble any manipulator at the simulator, the bounding box camera has more freedom to state its pose. So, during the simulations, we set the camera's pose to ensure the spheres' visibility. Fig. \ref{fig:sim-cam} illustrates the visible image of the camera at each pose. In the first pose, the camera looks straight towards the spheres (Fig. \ref{fig:sim_cam_p1}). After the camera moves down and left, looking upwards (Fig. \ref{fig:sim_cam_p2}), and finally, the camera moves up and right to the first pose, looking downwards (Fig. \ref{fig:sim_cam_p3}). This composition of the camera was kept for both experiments in the simulator. 

In the third essay, a realistic testbed was deployed to experiment with the algorithm in near-real-world conditions at the laboratory (Fig. \ref{fig:testbed-lab}). The testbed is composed of realistic artificial leaves and realistic plastic tomatoes. The tomatoes were hung in the testbed between the leaves. For baselining the tomatoes' position in the testbed, we relied on the manipulator's kinematics. For each tomato in the testbed, we moved the manipulator end-effector until the fruit and retrieved the end-effector's position. This will be the tomato position to the manipulator's base frame. Similarly to the essays in simulation, the bounding box camera at the simulator moved to three fixed poses that always assured the visibility of the tomatoes. A similar scheme to the one used before was set concerning the several limitations of the manipulator's manoeuvrability that made it difficult to set some poses to the camera. Unlike the simulation essays, several experiments were performed in the testbed. In the different experiments, we considered between one to three tomatoes being localised simultaneously, summing up to ten tomatoes and sixty measures. Fig. \ref{fig:oak-poses} illustrates the tomatoes' visibility by the OAK camera at each selected pose, for the different experiments. The rows represent the different considered experiments and each image has the tomatoes being localised simultaneously.

\begin{figure}[!htb]
    \centering\hfill
    \begin{subfigure}[b]{0.25\textwidth}
         \centering
         \includegraphics[width=\textwidth]{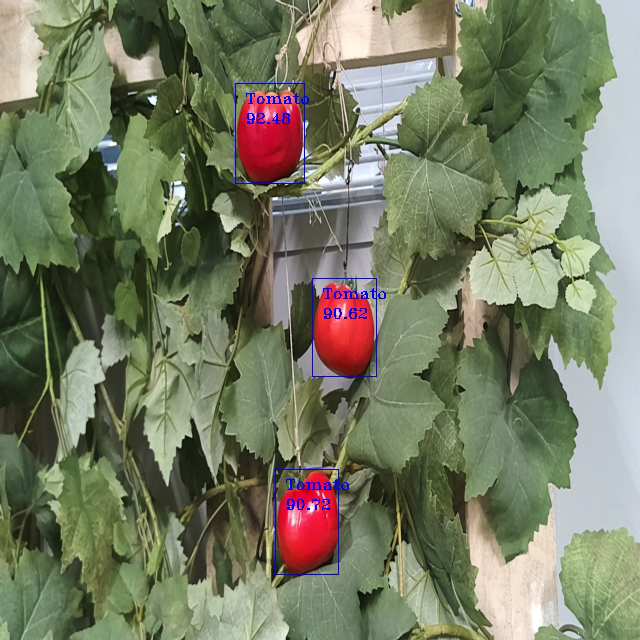}
         \caption{}
         \label{fig:oak-pose-a-exp1}
    \end{subfigure}\hfill
    \begin{subfigure}[b]{0.25\textwidth}
         \centering
         \includegraphics[width=\textwidth]{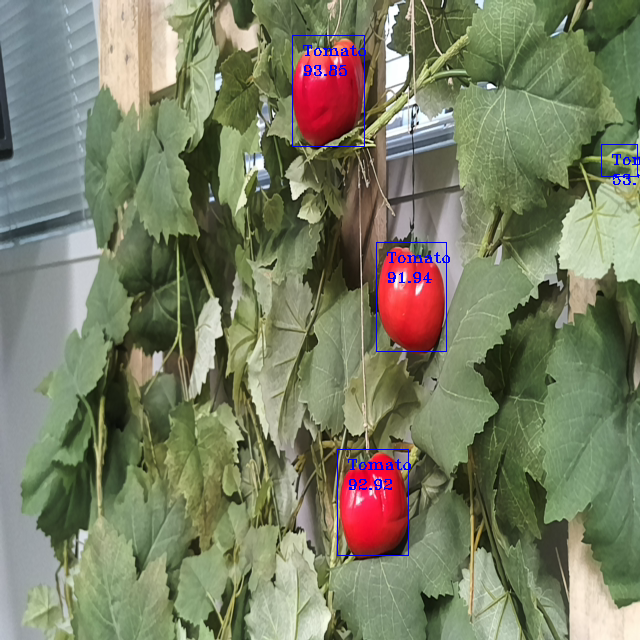}
         \caption{}
         \label{fig:oak-pose-b-exp1}
    \end{subfigure}\hfill
    \begin{subfigure}[b]{0.25\textwidth}
         \centering
         \includegraphics[width=\textwidth]{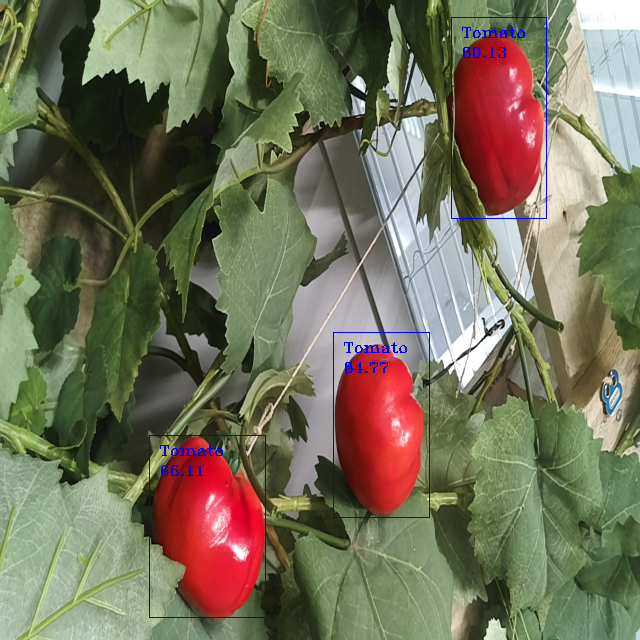}
         \caption{}
         \label{fig:oak-pose-c-exp1}
    \end{subfigure}\hfill\hfill\newline\hfill\\\hfill
    \begin{subfigure}[b]{0.25\textwidth}
         \centering
         \includegraphics[width=\textwidth]{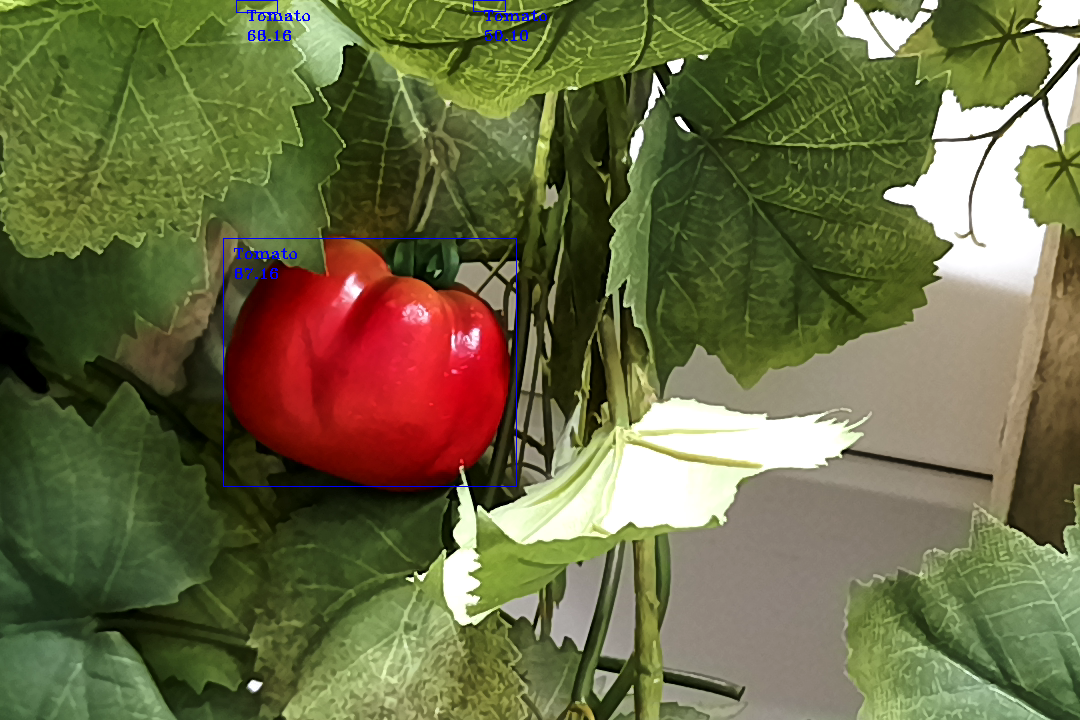}
         \caption{}
         \label{fig:oak-pose-a-exp2}
    \end{subfigure}\hfill
    \begin{subfigure}[b]{0.25\textwidth}
         \centering
         \includegraphics[width=\textwidth]{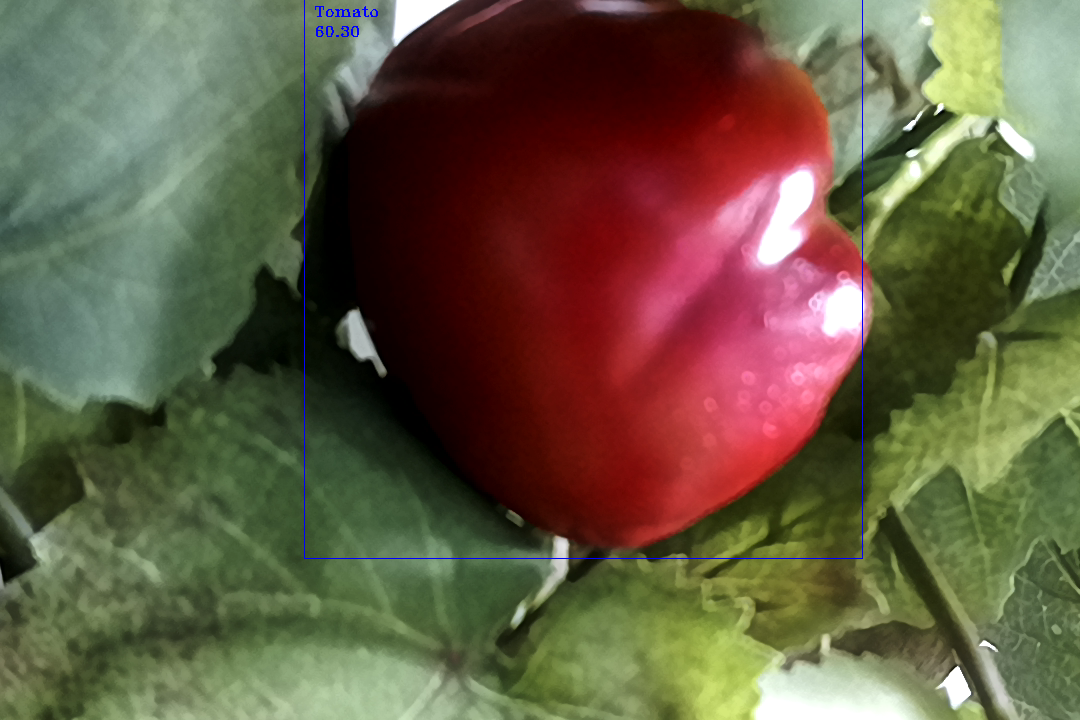}
         \caption{}
         \label{fig:oak-pose-b-exp2}
    \end{subfigure}\hfill
    \begin{subfigure}[b]{0.25\textwidth}
         \centering
         \includegraphics[width=\textwidth]{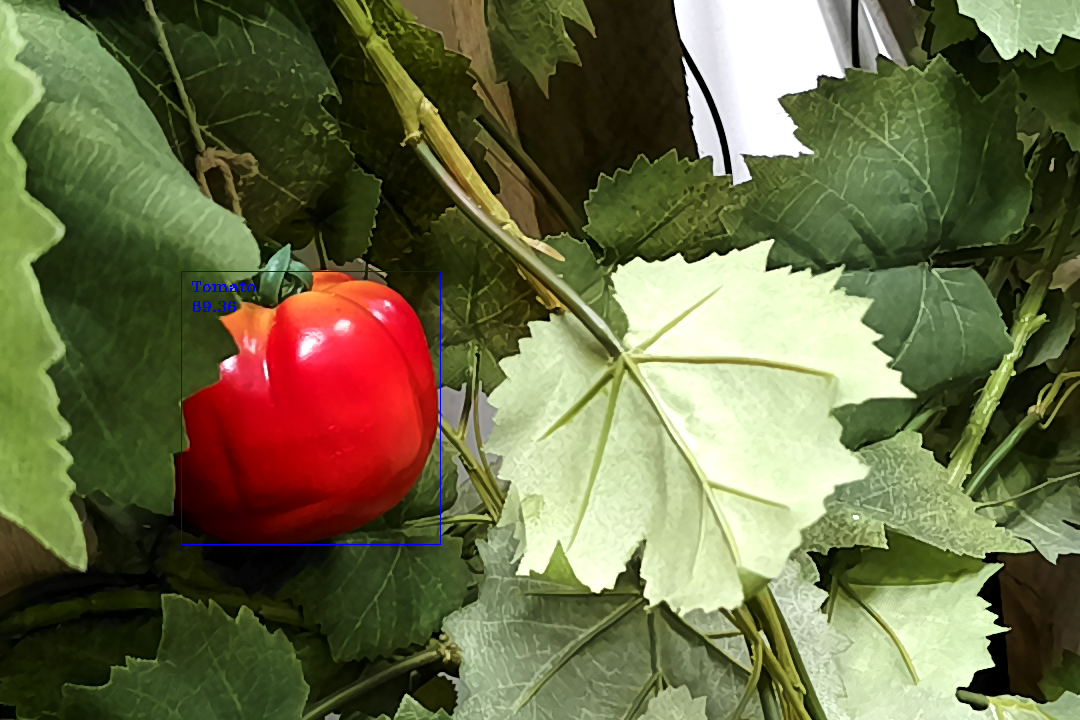}
         \caption{}
         \label{fig:oak-pose-c-exp2}
    \end{subfigure}\hfill\hfill\newline\hfill\\\hfill
    \begin{subfigure}[b]{0.25\textwidth}
         \centering
         \includegraphics[width=\textwidth]{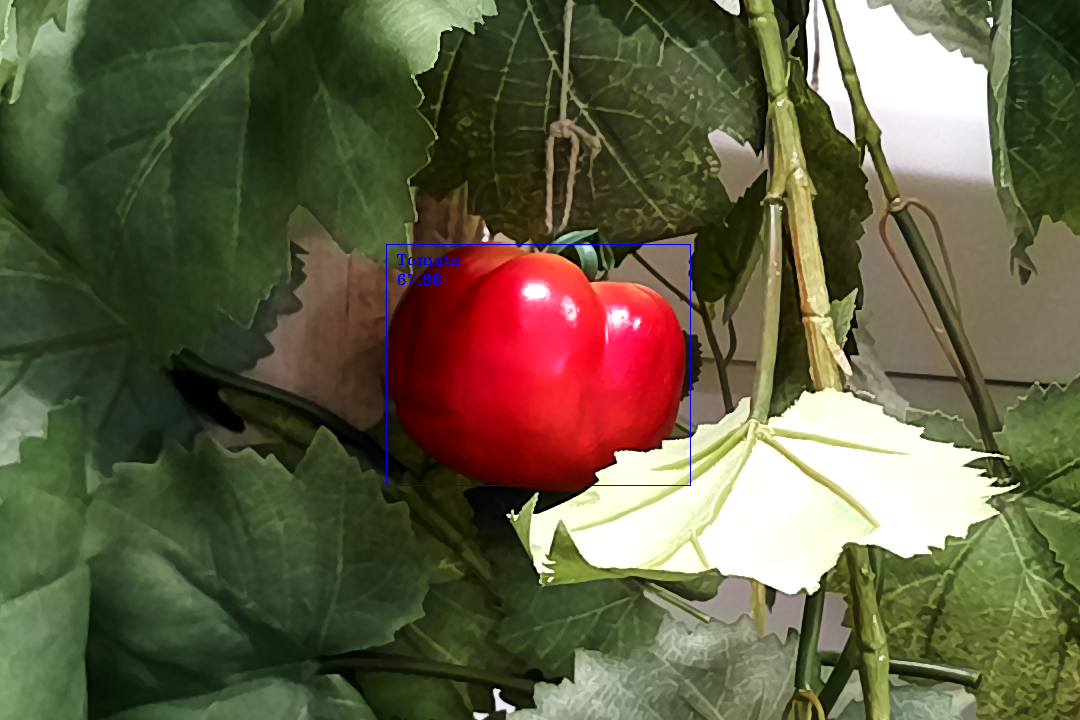}
         \caption{}
         \label{fig:oak-pose-a-exp4}
    \end{subfigure}\hfill
    \begin{subfigure}[b]{0.25\textwidth}
         \centering
         \includegraphics[width=\textwidth]{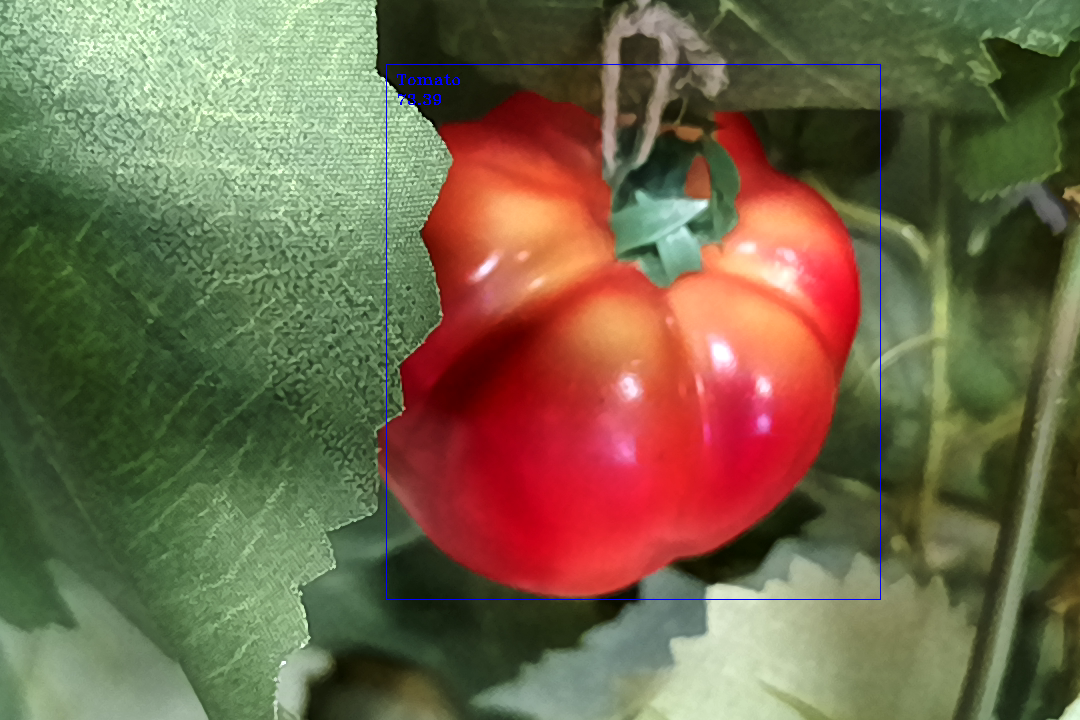}
         \caption{}
         \label{fig:oak-pose-b-exp4}
    \end{subfigure}\hfill
    \begin{subfigure}[b]{0.25\textwidth}
         \centering
         \includegraphics[width=\textwidth]{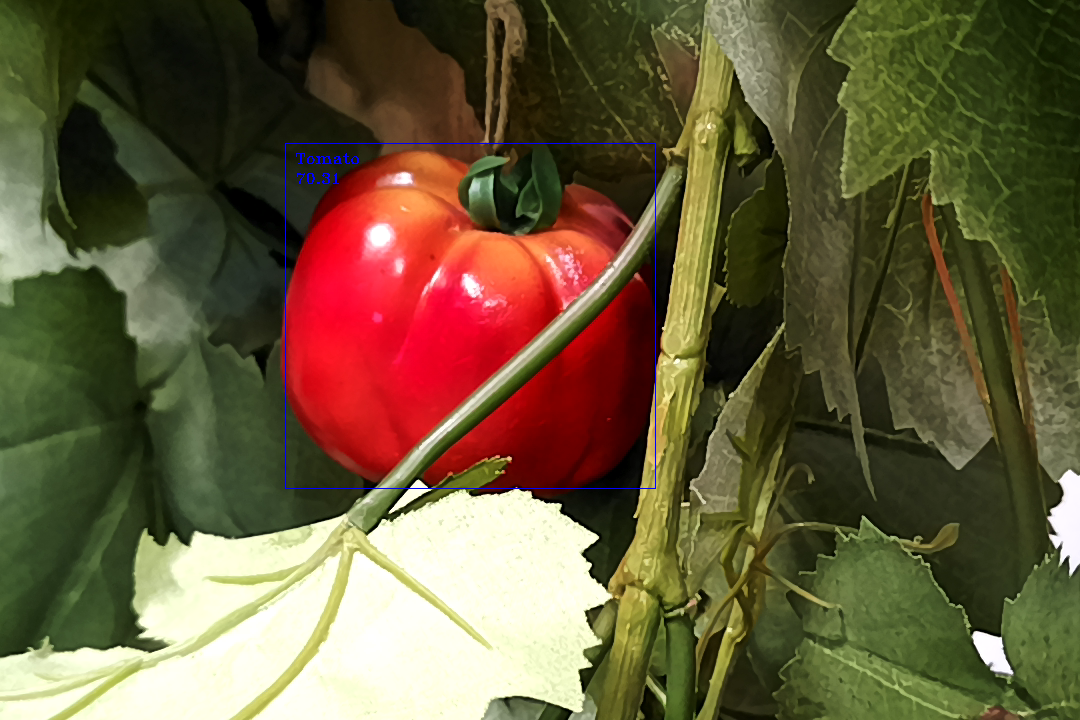}
         \caption{}
         \label{fig:oak-pose-c-exp4}
    \end{subfigure}\hfill\hfill\newline\hfill\\\hfill
    \begin{subfigure}[b]{0.25\textwidth}
         \centering
         \includegraphics[width=\textwidth]{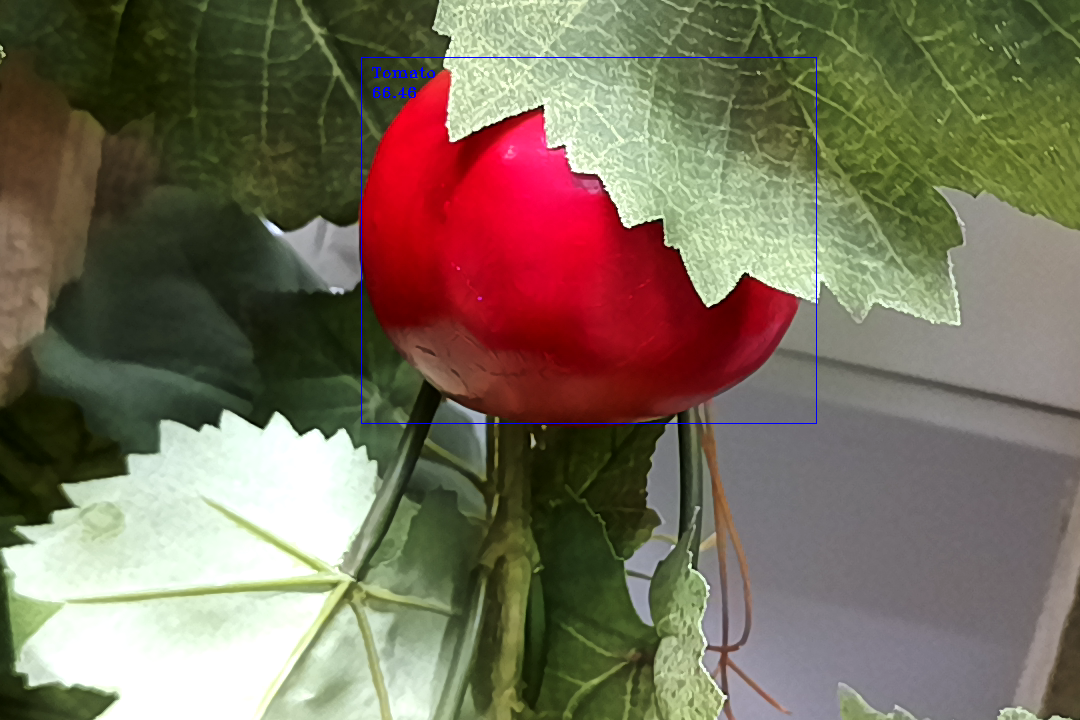}
         \caption{}
         \label{fig:oak-pose-a-exp5}
    \end{subfigure}\hfill
    \begin{subfigure}[b]{0.25\textwidth}
         \centering
         \includegraphics[width=\textwidth]{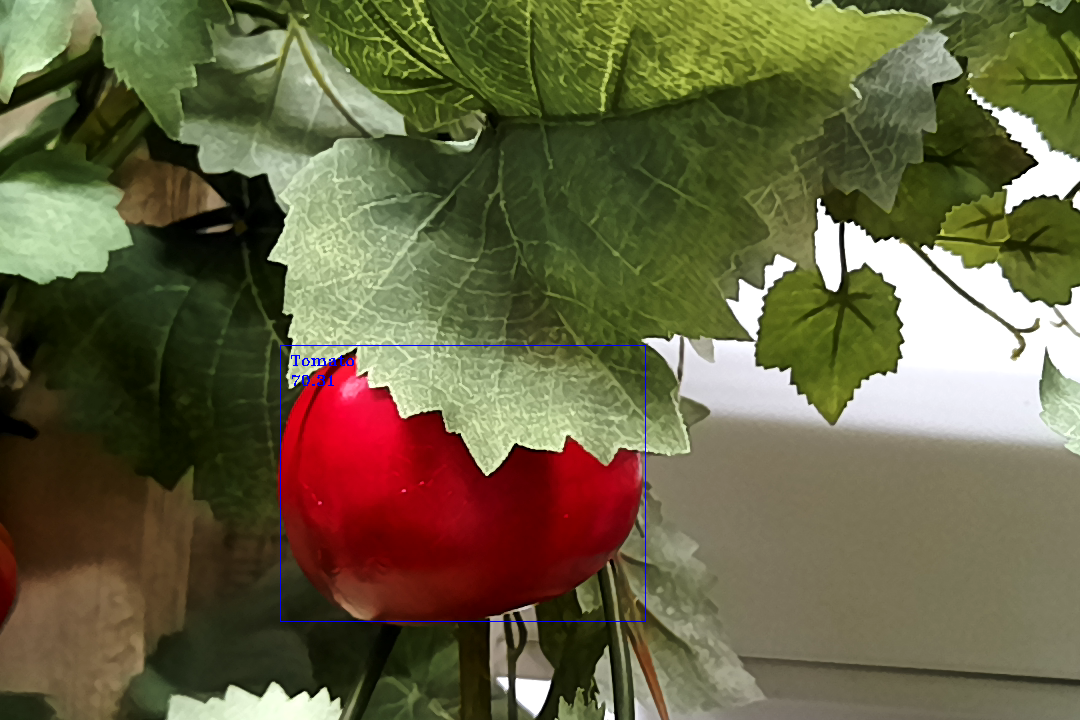}
         \caption{}
         \label{fig:oak-pose-b-exp5}
    \end{subfigure}\hfill
    \begin{subfigure}[b]{0.25\textwidth}
         \centering
         \includegraphics[width=\textwidth]{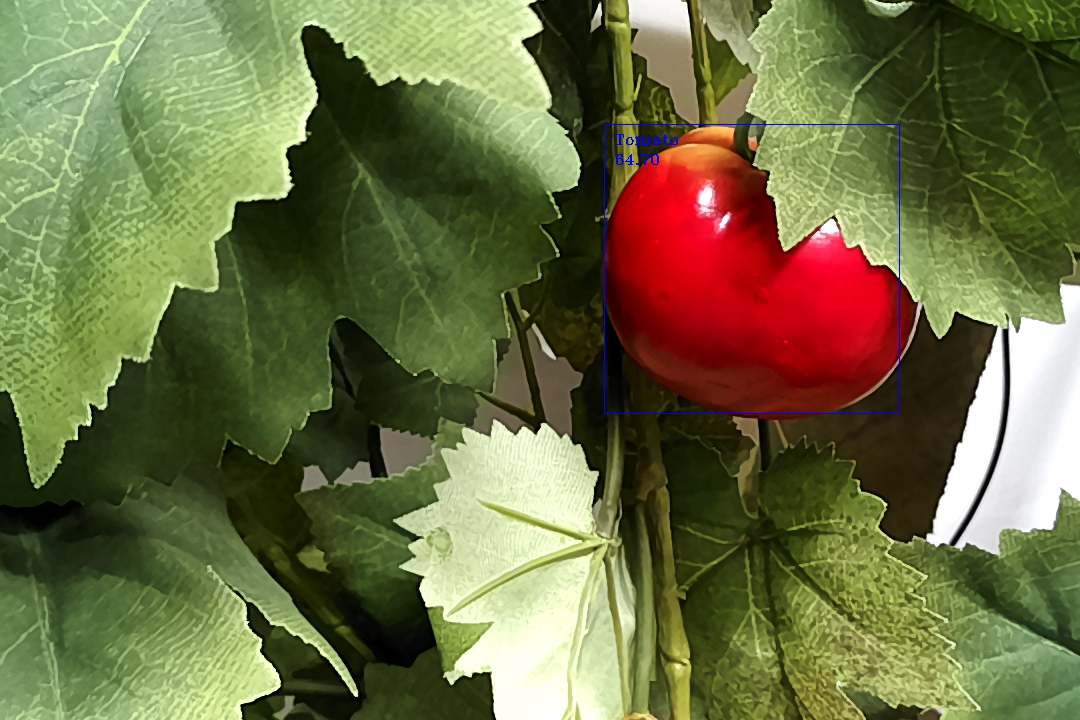}
         \caption{}
         \label{fig:oak-pose-c-exp5}
    \end{subfigure}\hfill\hfill\newline\hfill\\\hfill
    \begin{subfigure}[b]{0.25\textwidth}
         \centering
         \includegraphics[width=\textwidth]{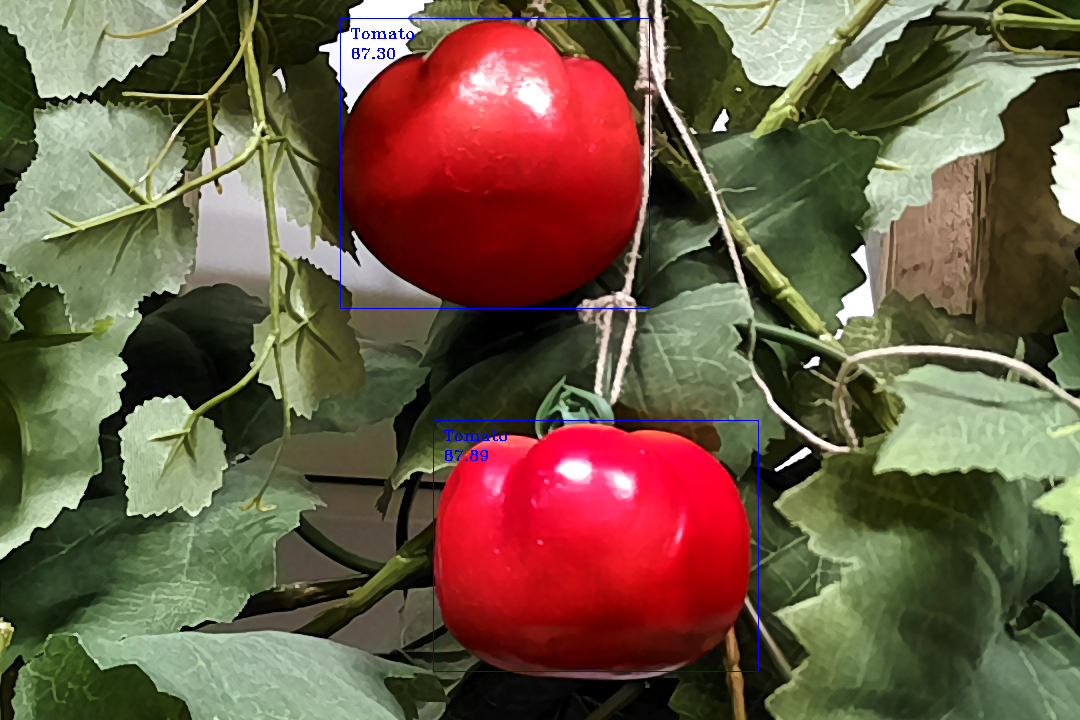}
         \caption{}
         \label{fig:oak-pose-a-exp7}
    \end{subfigure}\hfill
    \begin{subfigure}[b]{0.25\textwidth}
         \centering
         \includegraphics[width=\textwidth]{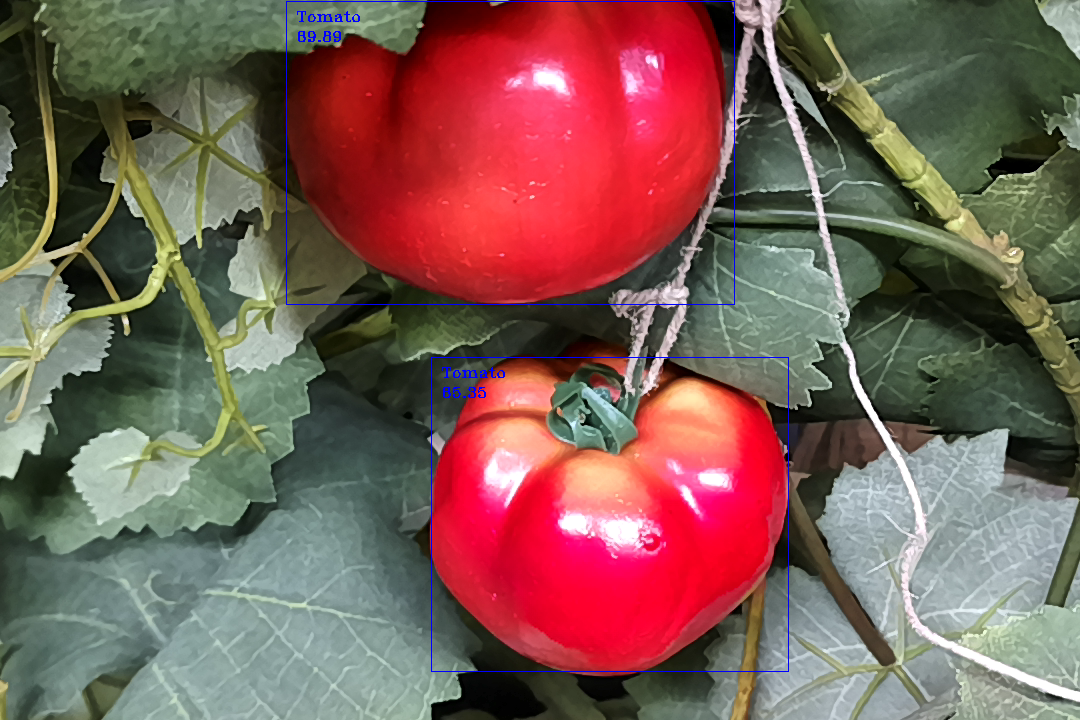}
         \caption{}
         \label{fig:oak-pose-b-exp7}
    \end{subfigure}\hfill
    \begin{subfigure}[b]{0.25\textwidth}
         \centering
         \includegraphics[width=\textwidth]{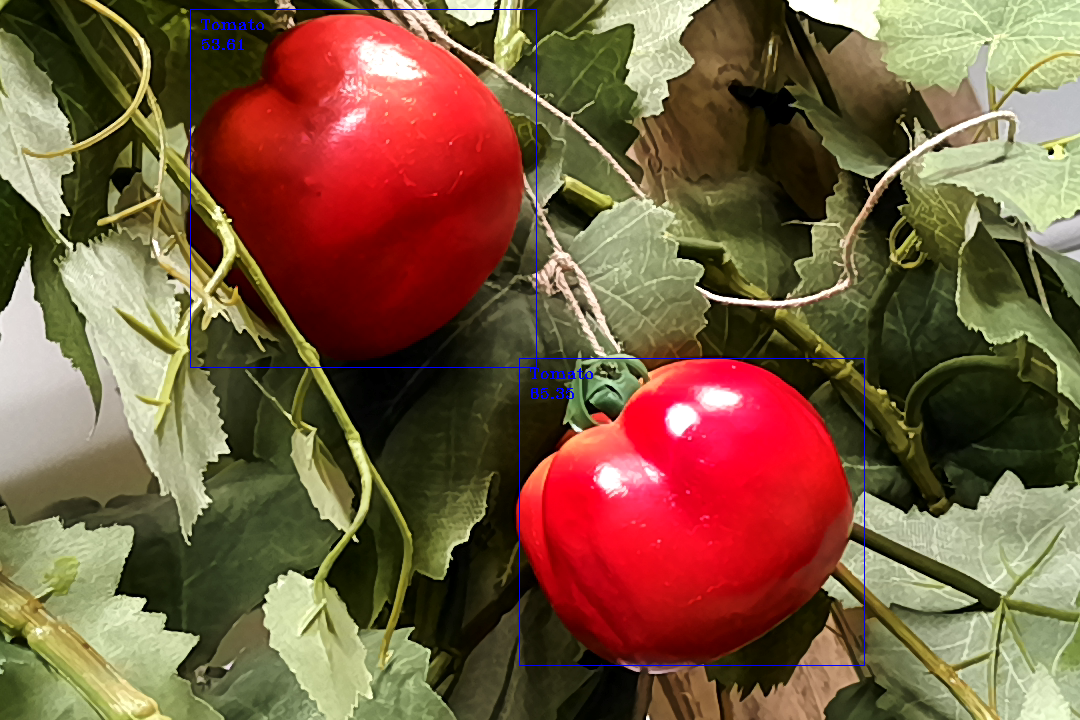}
         \caption{}
         \label{fig:oak-pose-c-exp7}
    \end{subfigure}\hfill\hfill\newline\hfill\\\hfill
\end{figure}
\begin{figure}\ContinuedFloat
    \centering\hfill
    \begin{subfigure}[b]{0.25\textwidth}
         \centering
         \includegraphics[width=\textwidth]{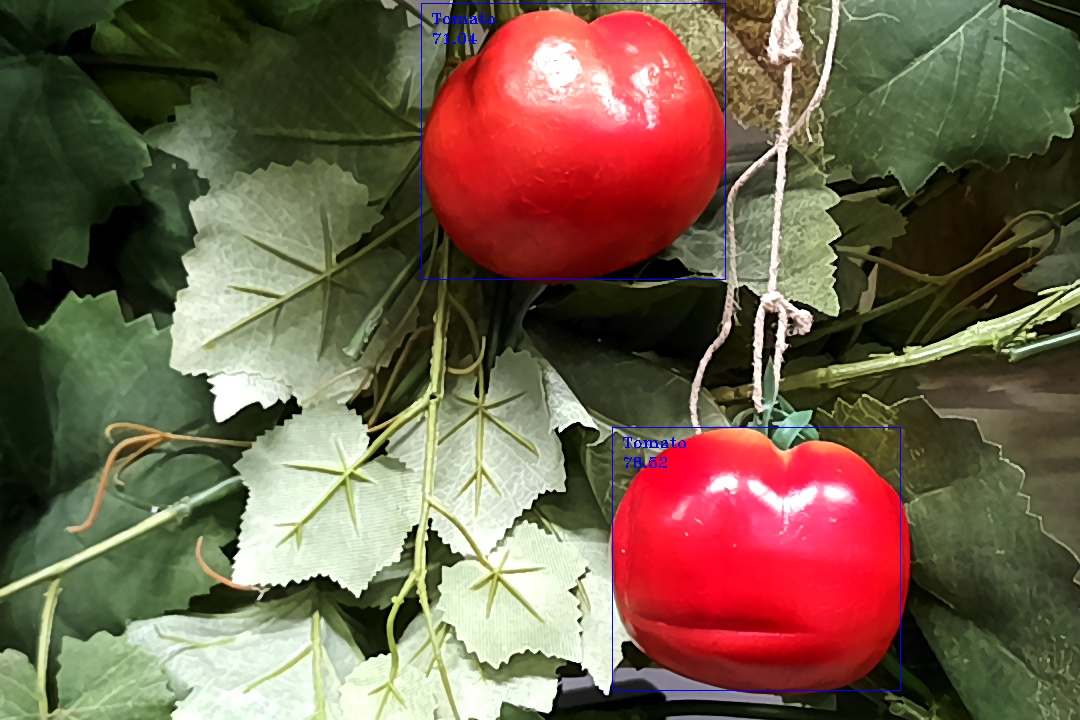}
         \caption{}
         \label{fig:oak-pose-a-exp8}
    \end{subfigure}\hfill
    \begin{subfigure}[b]{0.25\textwidth}
         \centering
         \includegraphics[width=\textwidth]{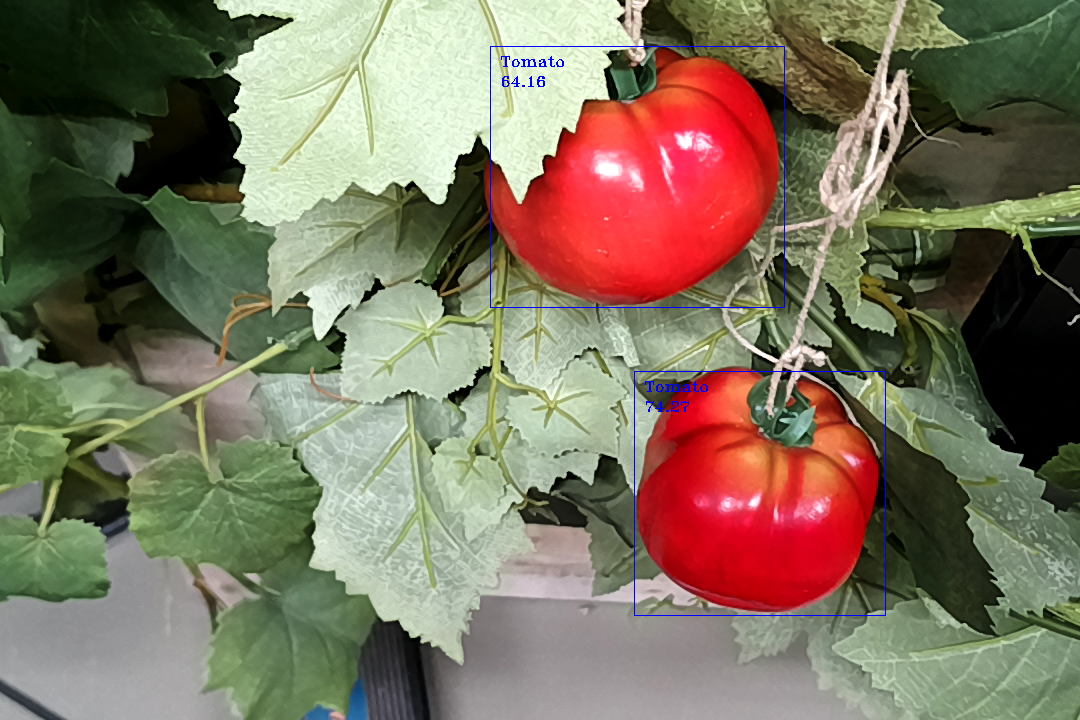}
         \caption{}
         \label{fig:oak-pose-b-exp8}
    \end{subfigure}\hfill
    \begin{subfigure}[b]{0.25\textwidth}
         \centering
         \includegraphics[width=0.66\textwidth]{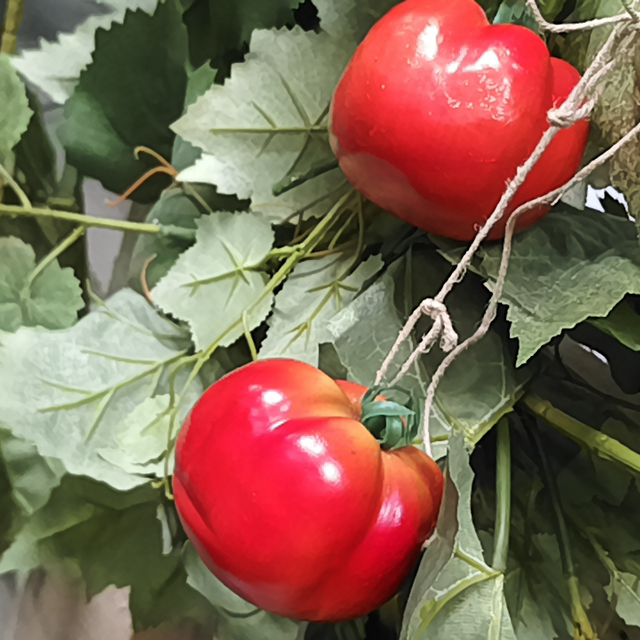}
         \caption{}
         \label{fig:oak-pose-c-exp8}
    \end{subfigure}\hfill\hfill
    \caption{View of the tomatoes in the testbed at each pose of the OAK-1 camera. {The blue squares around the tomatoes are the detected tomatoes by the bounding box camera OAK-1 using a custom-trained YOLO v8 tiny detector. Inside each bounding box are the detected class (tomato) and the detection confidence. Each row is an experiment, in a total of six experiments, and each figure contains the number of tomatoes being detected.}}
    \label{fig:oak-poses}
\end{figure}

To better assess the performance of the histogram filter to estimate the position of objects, we extracted some error metrics, namely the mean absolute error \eqref{eq: mae}, the mean square error \eqref{eq: mse}, the root mean square error or standard deviation \eqref{eq: rmse}, and the mean absolute percentage error \eqref{eq: mape}. In these equations, $\mu_j$ is the real centre of the object for the cluster $S_j$, and $\hat{\mu}_j$ is the estimated one using the previous methods, given the cluster $j$ until the maximum of clusters $M$.

\begin{align}
    \text{MAE }(\mu_j,\hat{\mu}_j) = & \dfrac{1}{N\cdot M} \sum_i^N \sum_j^M |\mu_{ij} - \hat{\mu}_{ij}| \qquad \forall j \in \mathds{N}:\{1 .. M\}\label{eq: mae} \\
    \text{MSE }(\mu_j,\hat{\mu}_j) = & \dfrac{1}{N\cdot M} \sum_i^N \sum_j^M (\mu_{ij} - \hat{\mu}_{ij})^2 \qquad \forall j \in \mathds{N}:\{1 .. M\} \label{eq: mse} \\
    \text{RMSE }(\mu_j,\hat{\mu}_j) = & \sqrt{ \dfrac{1}{N\cdot M} \sum_i^N \sum_j^M (\mu_{ij} - \hat{\mu}_{ij})^2} \qquad \forall j \in \mathds{N}:\{1 .. M\} \label{eq: rmse}  \\
    \text{MAPE }(\mu_j, \hat{\mu}_j) = & \dfrac{1}{N\cdot M} \sum_i^N \sum_j^M \left|\dfrac{\mu_{ij}-\hat{\mu}_{ij}}{\mu_{ij}}\right| \times 100 \qquad \forall j \in \mathds{N}:\{1 .. M\} \label{eq: mape}
\end{align}

\section{Results}
\label{sec: results}

As referred to in the previous section, we made three essays to validate the MonoVisual3DFilter's performance. Two essays happened in simulation, while the third was in a simulated testbed at the laboratory. To measure the performance, we recurred to different error metrics: \eqref{eq: mae}, \eqref{eq: mse}, \eqref{eq: rmse} and \eqref{eq: mape}. The manipulator and the bounding box camera were moved to fixed poses where all the fruits were always visible. All permutations between poses were considered for the essays, resulting in six estimations of each tomato for each experiment. The computed error results from the error of each estimated pose to ground truth pose of each corresponding tomato. 

\begin{figure}[!htb]
    \centering\hfill
    \begin{subfigure}[b]{0.24\textwidth}
         \centering
         \includegraphics[width=\textwidth]{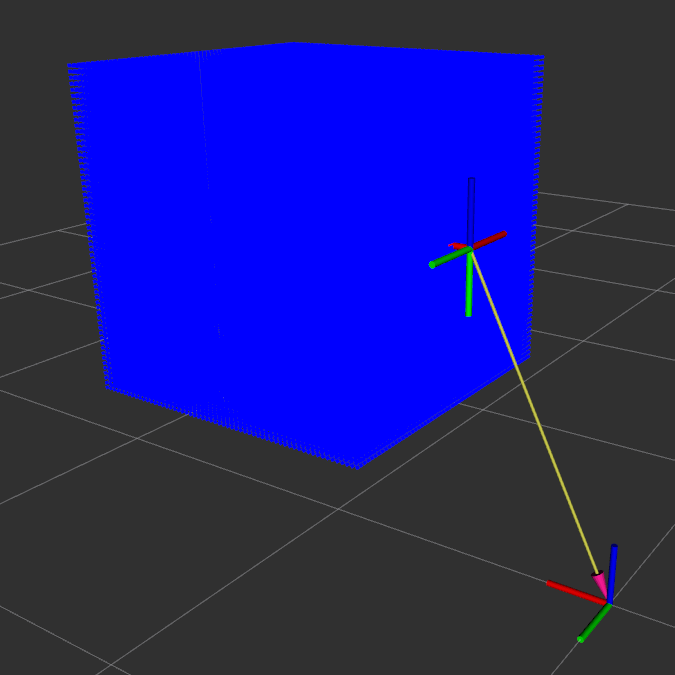}
         \caption{}
         \label{fig:decompose}
    \end{subfigure}\hfill
    \begin{subfigure}[b]{0.24\textwidth}
         \centering
         \includegraphics[width=\textwidth]{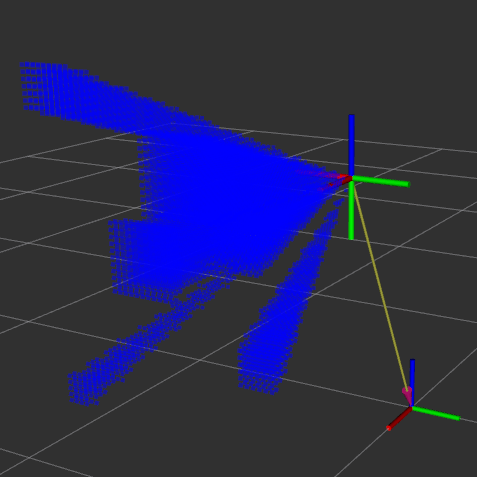}
         \caption{}
         \label{fig:sim_sq_1}
    \end{subfigure}\hfill
    \begin{subfigure}[b]{0.24\textwidth}
         \centering
         \includegraphics[width=\textwidth]{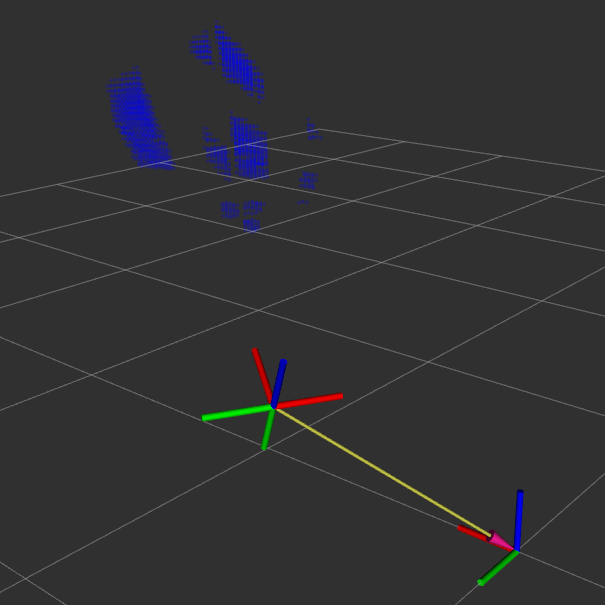}
         \caption{}
         \label{fig:sim_sq_2}
    \end{subfigure}\hfill
    \begin{subfigure}[b]{0.24\textwidth}
         \centering
         \includegraphics[width=\textwidth]{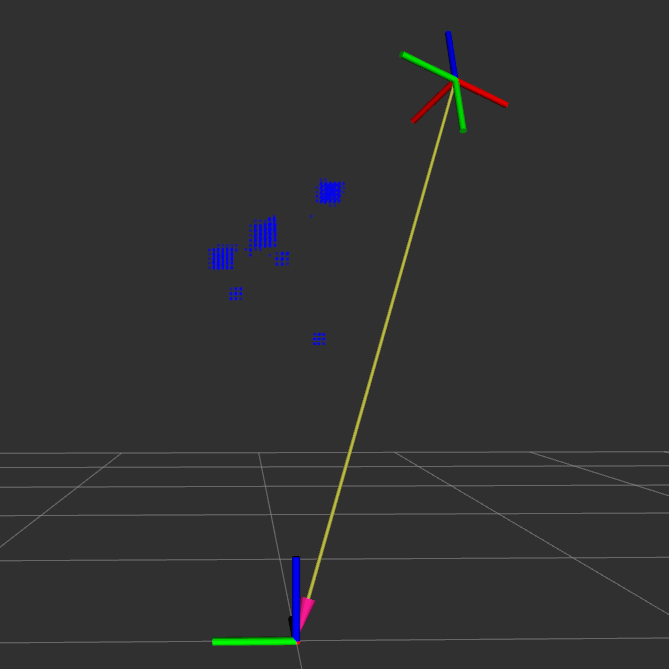}
         \caption{}
         \label{fig:sim_sq_3}
    \end{subfigure}\hfill\hfill
    \caption{{Iteration of the Histogram filter during simulation for detecting the six spheres, considering a square kernel. (a) decomposition of the state space at the beginning of the algorithm; (b) detection at the end of the first viewpoint; (c) detection at the end of the second viewpoint; (d) detection at the end of the third viewpoint.}}
    \label{fig:sim-sq}
\end{figure}

\begin{figure}[!htb]
    \centering\hfill
    \begin{subfigure}[b]{0.24\textwidth}
         \centering
         \includegraphics[width=\textwidth]{decompose.png}
         \caption{}
         \label{fig:decompose-1}
    \end{subfigure}\hfill
    \begin{subfigure}[b]{0.24\textwidth}
         \centering
         \includegraphics[width=\textwidth]{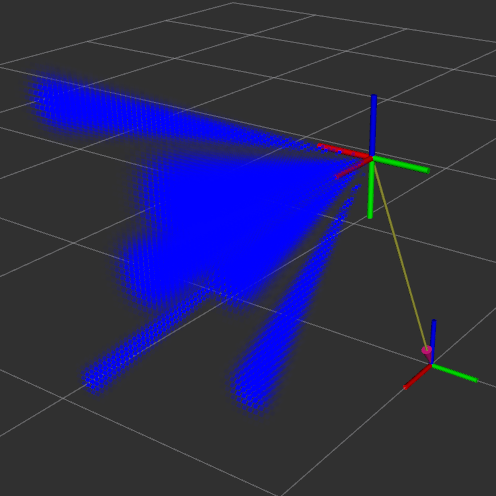}
         \caption{}
         \label{fig:sim_gauss_1}
    \end{subfigure}\hfill
    \begin{subfigure}[b]{0.24\textwidth}
         \centering
         \includegraphics[width=\textwidth]{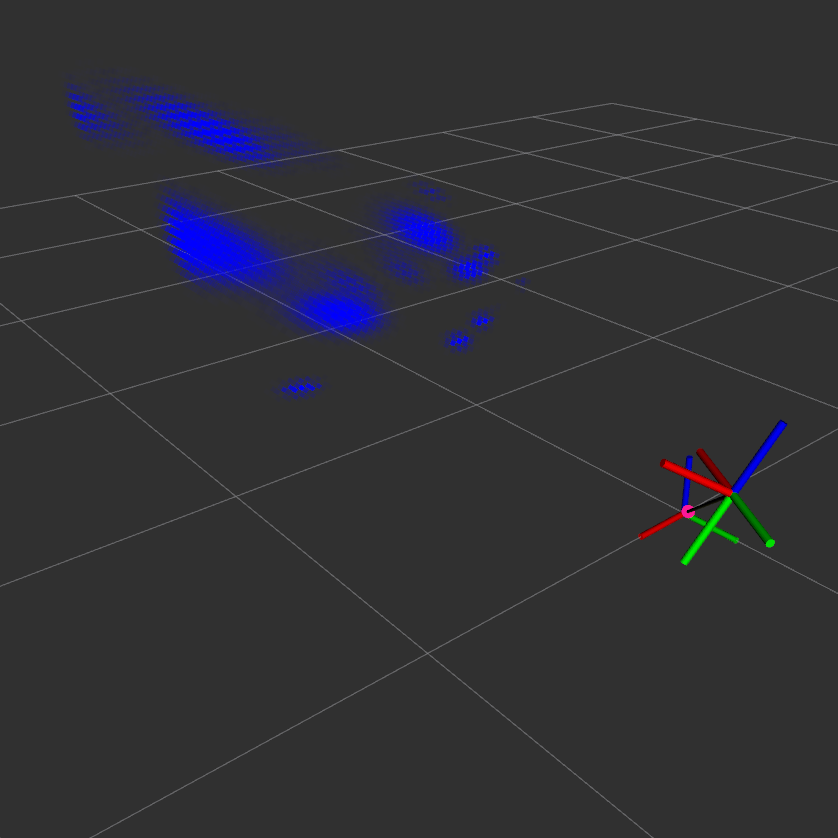}
         \caption{}
         \label{fig:sim_gauss_2}
    \end{subfigure}\hfill
    \begin{subfigure}[b]{0.24\textwidth}
         \centering
         \includegraphics[width=\textwidth]{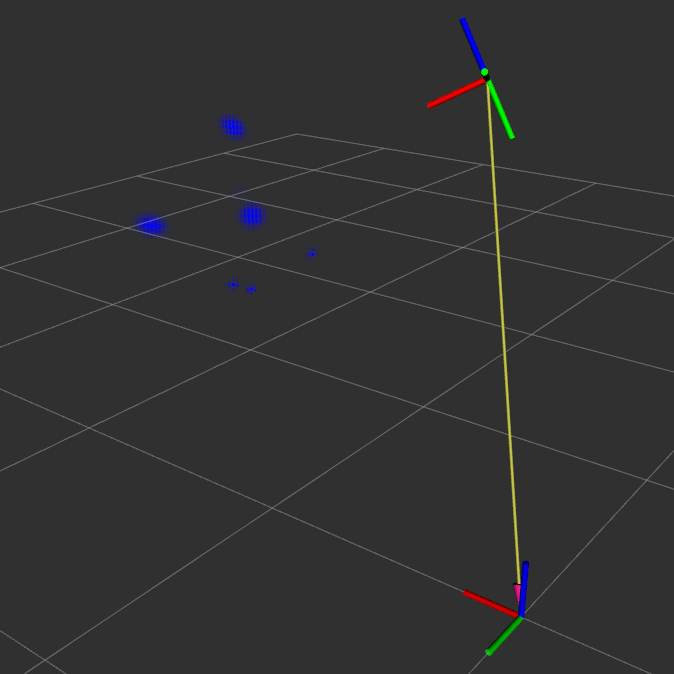}
         \caption{}
         \label{fig:sim_gauss_3}
    \end{subfigure}\hfill\hfill
    \caption{{Iteration of the Histogram filter during simulation for detecting the six spheres, considering a Gaussian kernel, $\mathcal{N}(0, 0.2)$. (a) decomposition of the state space at the beginning of the algorithm; (b) detection at the end of the first viewpoint; (c) detection at the end of the second viewpoint; (d) detection at the end of the third viewpoint.}}
    \label{fig:sim-gauss}
\end{figure}

\begin{figure}[!htb]
    \centering
    \includegraphics[width=0.5\textwidth]{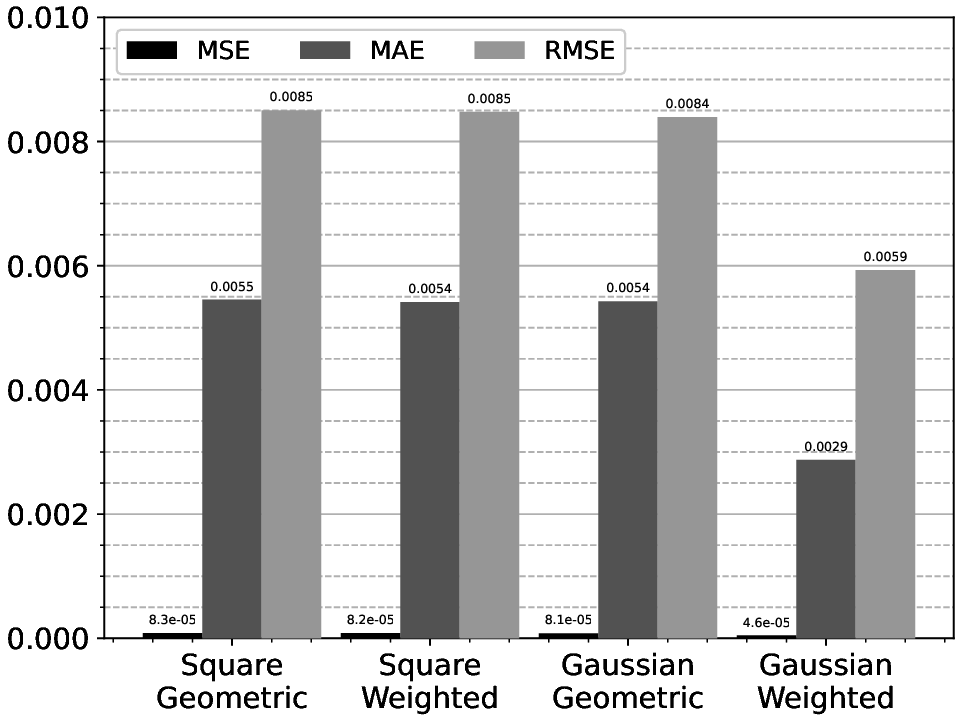}
    \caption{Error in estimating the position of the spheres in simulation without noise}
    \label{fig:nonoise-plot}
\end{figure}

In the first essay, we used the simulation to estimate the \ac{3D} position of the green spheres (Fig. \ref{fig:simulation}) without any noise. During the execution of the histogram filter, the bounding box camera moved to different poses to intersect and be these positions. At each pose, the state space is operated to remove or smooth the existence of objects in each state according to the used kernel. Two kinds of kernels were considered, the square kernel (Fig. \ref{fig:sim-sq}) and the Gaussian kernel (Fig. \ref{fig:sim-gauss}). Visually, both kernels performed similarly; however, the Gaussian kernel has two hyperparameters that deliver more freedom to set the kernel's size and estimate the objects' position, allowing the Gaussian filter to be more aggressive or smooth. We considered a two-dimensional Gaussian Filter with $\mathcal{N}(0, \text{size}/2)$ for the current experiment, where size is the width or the height of the detected object. The plot of Fig. \ref{fig:nonoise-plot} illustrates the error using a square or Gaussian kernel and the geometric centres from the k-means algorithm or the weighted centre resulting from applying the computed weights of the histogram filter. For this essay, using a Gaussian kernel with a weighted centre estimation was advantageous. 

\begin{figure}[!htb]
    \centering
    \includegraphics[width=0.5\textwidth]{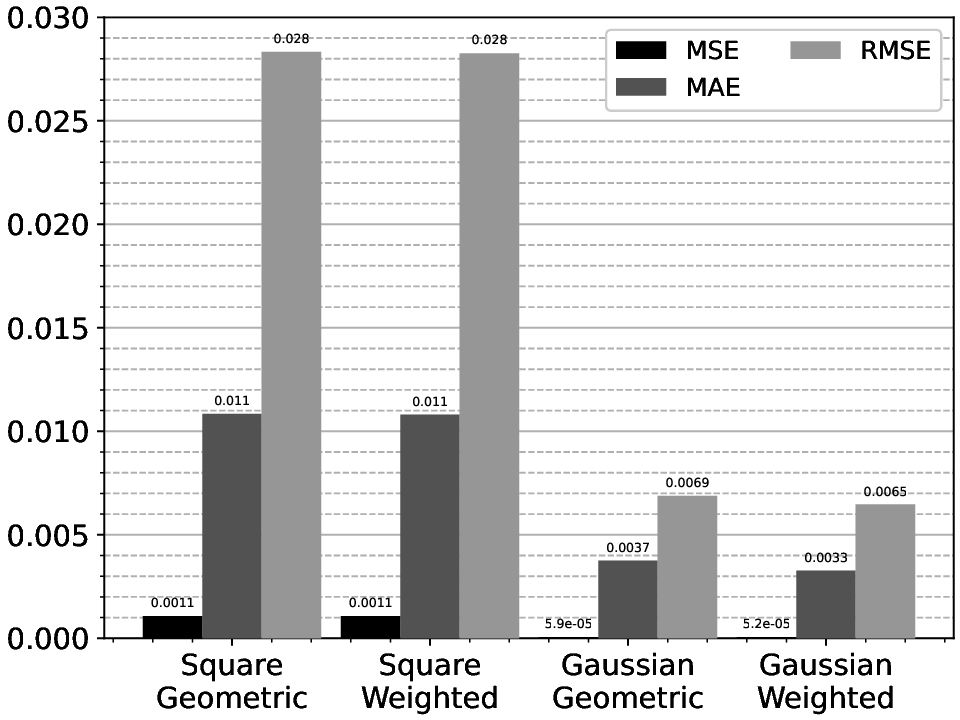}
    \caption{Error in estimating the position of the spheres in simulation with noise}
    \label{fig:noise-plot}
\end{figure}

Adding some noise to the simulation by moving the bounding box's centre and size and deleting it randomly, we get the results of Fig. \ref{fig:noise-plot}. For moving the bounding box's centre and dimensions, we considered a Gaussian profile with $\mathcal{N}(0, 0.05)$. The bounding box will likely fail of \SI{2}{\percent}. Once again, the Gaussian kernel with a weighted estimation of the spheres' centre performed better. Besides, in this experiment, the Gaussian kernel was more robust and accurate than the square kernel. The behaviour of the histogram filter in each pose is similar to the previous essay, but now, the algorithm has reported more noise. We set the Gaussian kernel such as $\mathcal{N}(0, size/3)$ to overcome this effect and have a more stable filter. This is the smallest value that assures that any fruit is lost and the point cloud is not too sparse, intersecting with the cloud of other fruits.

\begin{figure}[!htb]
    \centering
    \includegraphics[width=0.55\textwidth]{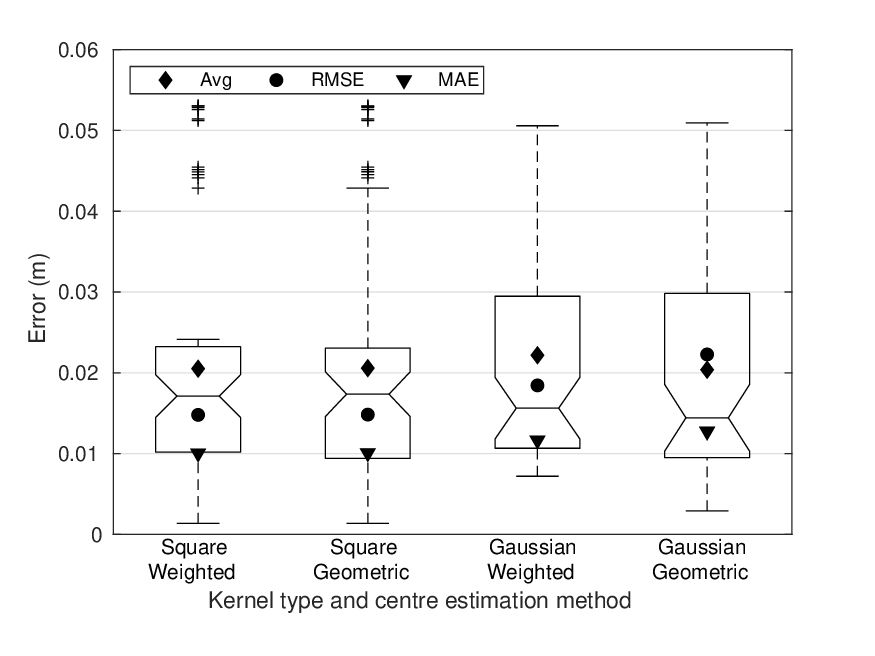}
    \caption{Error in estimating the position of the tomatoes at the testbed}
    \label{fig:testbed-plot}
\end{figure}

Given the overall success of the essays with a mean absolute error smaller than \SI{1}{\centi\metre}, we essayed the algorithm in a testbed at the laboratory. Fig. \ref{fig:testbed-plot} illustrates the box plot error of the MonoVisual3DFilter in the testbed, using the Robotis Manipulator-H and the OAK-1 camera. A descriptive statement of the error is made in the table \ref{tab: testbed error}. In this analysis, we also considered the average Euclidean error distance and the MAPE to better assess the feasibility of the MonoVisual3DFilter. In all the camera poses, all the tomatoes were always visible (Fig. \ref{fig:oak-poses}). For this essay, we considered six experiments, that aimed to estimate the position between one to three tomatoes simultaneously, as stated in Fig. \ref{fig:oak-poses}, and summing up to sixty estimated measures for the position of the tomatoes.
Against the expected, Gaussian kernels performed worse than square kernels, and the results of weighted and geometric centres are identical, despite the weighted method tending to have a lower error for the same standard deviation. 

\begin{table}[!htb]
\caption{Error computations to the testbed experiments for the different kernels and centre estimation methods}
\label{tab: testbed error}
\centering
\begin{tabular}{@{}lcccc@{}}
\toprule
 &
  \textbf{\begin{tabular}[c]{@{}c@{}}Square\\ Weighted\end{tabular}} &
  \textbf{\begin{tabular}[c]{@{}c@{}}Square\\ Geometric\end{tabular}} &
  \textbf{\begin{tabular}[c]{@{}c@{}}Gaussian\\ Weighted\end{tabular}} &
  \textbf{\begin{tabular}[c]{@{}c@{}}Gaussian\\ Geometric\end{tabular}} \\ \midrule
\textbf{Euclidean Error (Avg)} & \SI{0.0205}{\metre}    & \SI{0.0206}{\metre}    & \SI{0.0222}{\metre}     & \SI{0.0204}{\metre}    \\
\textbf{MSE}                   & $0.2187\times 10^{-3}$ & $0.2197\times 10^{-3}$ & $0.3399\times 10^{-3}$  & $0.4960\times 10^{-3}$ \\
\textbf{RMSE}                  & 0.0148                 & 0.0148                 & 0.0184                  & 0.0223                 \\
\textbf{MAE}                   & \SI{0.0100}{\metre}    & \SI{0.0101}{\metre}    & \SI{0.0116}{\metre}     & \SI{0.0127}{\metre}    \\
\textbf{MAPE}                  & \SI{63.52}{\percent}   & \SI{63.51}{\percent}   & \SI{57.35}{\percent} & \SI{74.15}{\percent}   \\ \bottomrule
\end{tabular}
\end{table}

\section{Discussion}
\label{sec: discussion}

The overall use of MonoVisual3DFilter for estimating the \ac{3D} position of tomato fruits looks effective. 

Under a simulated environment, the system always got a maximum error smaller than \SI{10}{\milli\metre}. Increasing the resolution of the discretised state space could leverage better system accuracy, but increase the processing time and memory usage. As made in the second experiment, adding noise to the system makes the importance of using smooth kernels visible. The square kernel's aggressive binary behaviour rejects some state space positions and can never be recovered. On the other hand, the Gaussian filter has a smooth behaviour, and the positions are iteratively removed according to their distance from the filter's centre. So, smooth kernels can recover some state space points since they are never completely rejected. Besides, the Gaussian filter also was more failure-prone than the square kernel because the last one failed many times to detect the fruits, forcing us to repeat some experiments. Because the positions near the centre of the tomato have a bigger score, using weighted centre estimation procedures allows for a better fit of the estimated centre to the real centre of the tomato, reducing the effect and deviations of sparse clouds. 

When moving the implemented algorithm to a real robot and camera in a testbed, we reported that, against expectations, the Gaussian kernels were not very effective and square kernels reached a higher accuracy. In this case, it is always irrelevant to consider geometric or weighted estimations of the centre. In the testbed, the system reported a mean absolute error around \SI{20}{\milli\metre}, but the error can evolute to near \SI{60}{\milli\metre}, that can compromise the use of the algorithm for more demanding tasks. However, it is still important to better identify the source of the error, whether it comes from the MonoVisual3DFilter or the ground-truth's baseline, which was badly fitted to the tomato centre. {Although the experiment was made with the viewpoints distanced by around \SI{0.5}{\metre} of the fruits, the error should improve if a closer assessment of the tomato position is made.} However, the reported error could not be critical whether using soft-grippers to aid harvesting tasks or using complementary algorithms. For tasks such as monitoring, this error could even be less relevant.

Additional experiments also allowed us to assess the importance of the selected viewpoints. Co-linear viewpoints do not allow for effective estimation of the position of the objects. However, normal viewpoints aim to better intersect the filter views and effectively estimate the position of the objects. This topic should be a concern under real-world and testbed experiments because we frequently use manipulators with several positioning and orienting constraints.

\begin{figure}[!htb]
    \centering
    \includegraphics[width=0.55\textwidth]{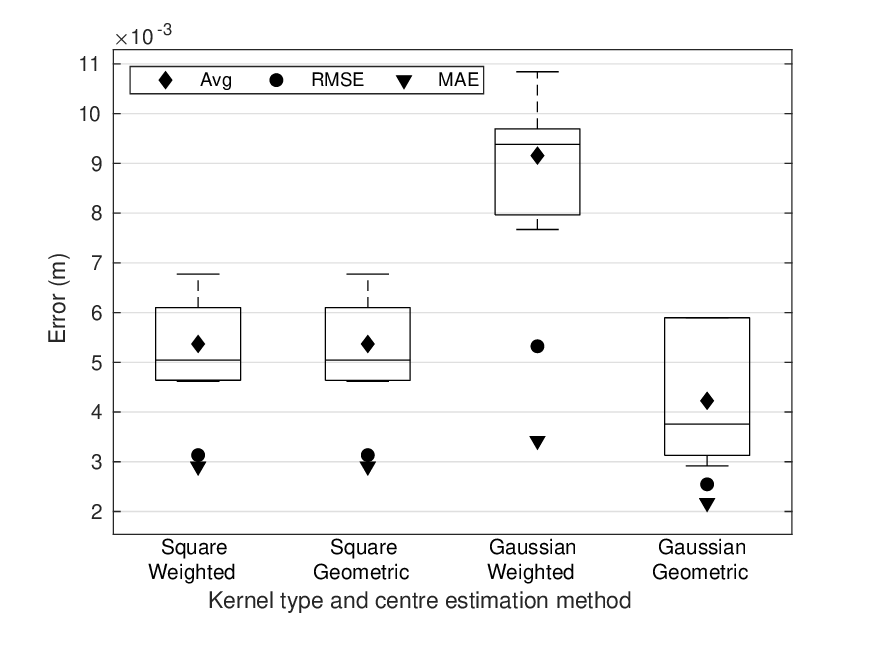}
    \caption{Error for the case of partial occlusion of Figs. \ref{fig:oak-pose-a-exp5}, \ref{fig:oak-pose-b-exp5}, \ref{fig:oak-pose-c-exp5},}
    \label{fig: error occludeede}
\end{figure}

Experiments as reported in Figs. \ref{fig:oak-pose-a-exp5}, \ref{fig:oak-pose-b-exp5}, \ref{fig:oak-pose-c-exp5}, and some others, could assess the feasibility of the MonoVisual3DFilter to partially occluded fruits. Firstly, due to the limitations of detection algorithms to detect completely occluded objects, the MonoVisual3DFilter do not work for these cases. Concerning partial occlusion, we verified that the MonoVisual3DFilter could lead with the occlusion and effectively estimate the position of the tomato (Fig. \ref{fig: error occludeede}). Going further, we can even state that the MonoVisual3DFilter is occlusion-independent and estimates the position of occluded objects so good as good is the object detector.

\begin{figure}[!htb]
    \centering\hfill
    \begin{subfigure}[b]{0.3\textwidth}
         \centering
         \includegraphics[width=\textwidth]{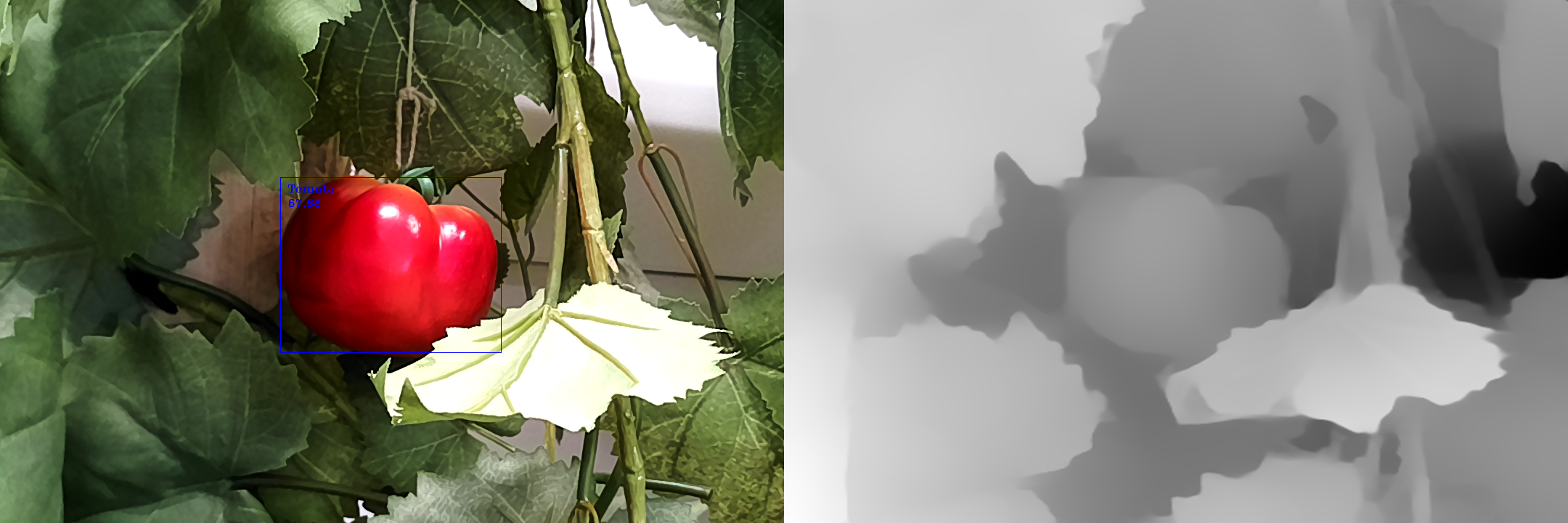}
         \caption{}
    \end{subfigure}\hfill
    \begin{subfigure}[b]{0.3\textwidth}
         \centering
         \includegraphics[width=\textwidth]{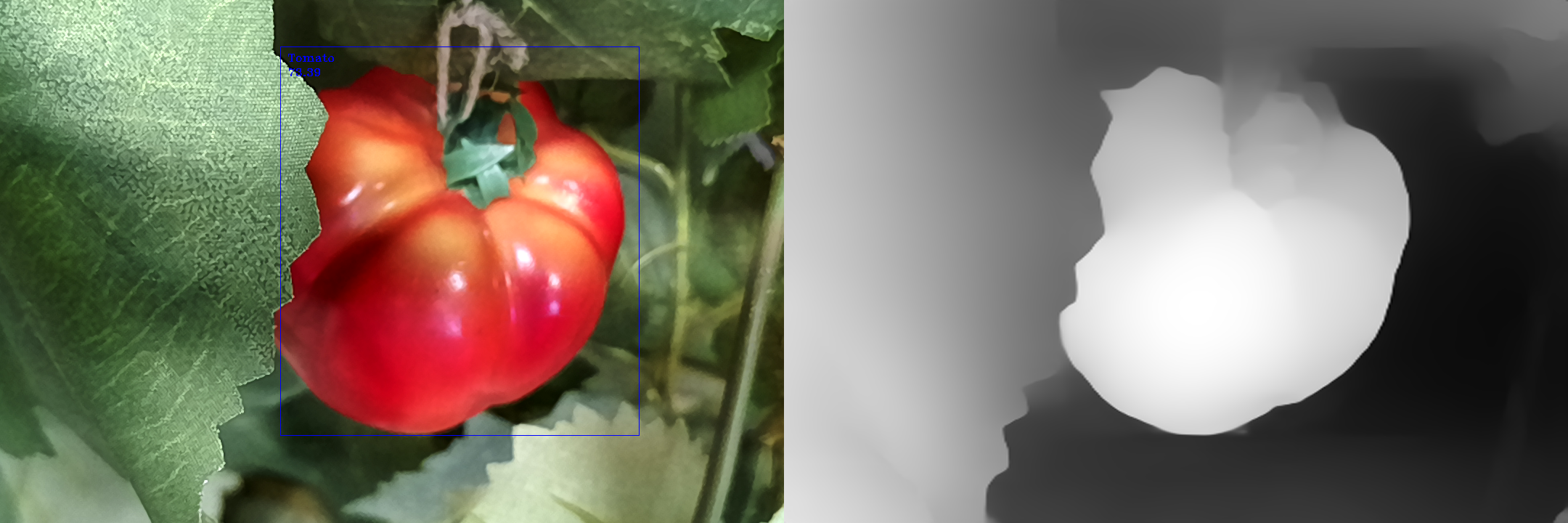}
         \caption{}
    \end{subfigure}\hfill
    \begin{subfigure}[b]{0.3\textwidth}
         \centering
         \includegraphics[width=\textwidth]{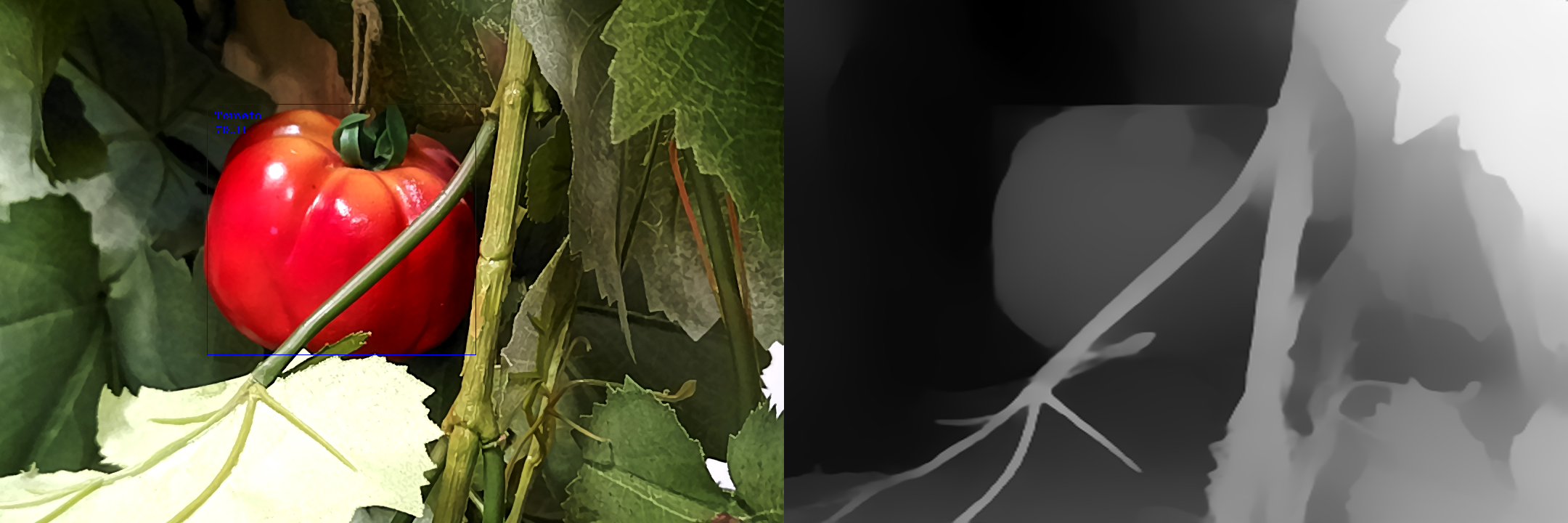}
         \caption{}
    \end{subfigure}\hfill\hfill
    \caption{RGB and Depth images from the MiDaS v3.1 DPT SWIN2 Large 384 for estimating the tomatoes' distance to the camera's sensor.}
    \label{fig: MiDaS figures}
\end{figure}

\begin{figure}[!htb]
    \centering\hfill
    \begin{subfigure}[b]{0.49\textwidth}
         \centering
         \includegraphics[width=\textwidth]{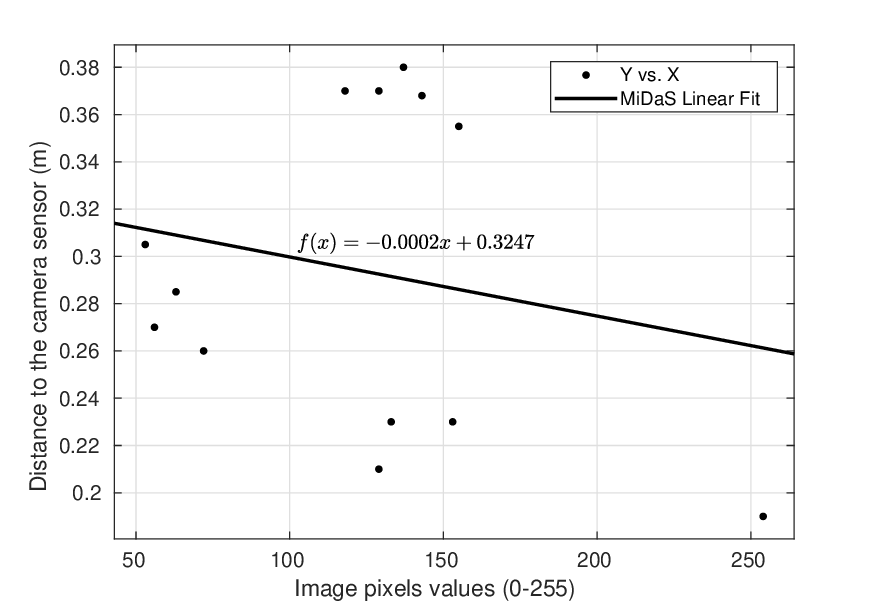}
         \caption{}
         \label{fig: MiDaS curve fitting}
    \end{subfigure}\hfill
    \begin{subfigure}[b]{0.49\textwidth}
         \centering
         \includegraphics[width=\textwidth]{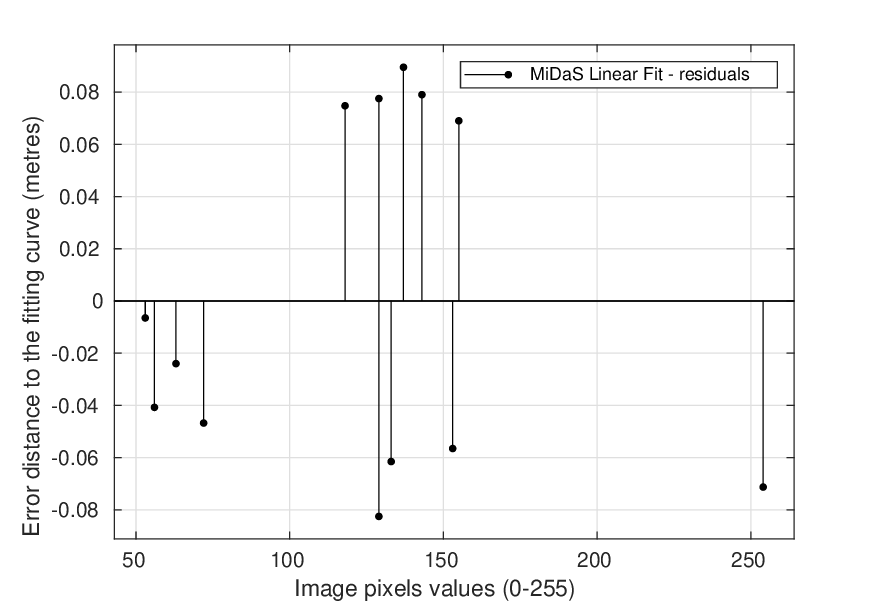}
         \caption{}
         \label{fig: MiDaS residuals}
    \end{subfigure}\hfill\hfill
    \caption{Calibration curve to estimate the absolute depth in metres to the camera sensor for MiDaS \ac{CNN}}
    \label{fig: MiDaS fitting}
\end{figure}

Observing the literature (section \ref{sec: introduction}), we can conclude that it tends to use deep learning solutions to infer the depth from monocular RGB cameras. A solution already available in the literature and that aims to effectively estimate the depth from images is the MiDaS network \cite{Ranftl2022,Birkl2023}. For comparison with the MonoVisual3DFilter, we used the MiDaS v3.1 DPT SWIN2 Large 384, and applied it to the images of the experiment composed by the Figs. \ref{fig:oak-pose-a-exp4}, \ref{fig:oak-pose-b-exp4}, and \ref{fig:oak-pose-c-exp4}. Fig. \ref{fig: MiDaS figures} reports the output of the MiDaS \ac{CNN} for the proposed images. Once the network reports a relative pose, a calibration is required to estimate the real depth to the camera. According to \cite{Ranftl2022}, the absolute depth can be computed through a linear regression curve. So, a rough calibration was performed and reported the curve of Fig. \ref{fig: MiDaS curve fitting}. As can be visually concluded from Fig. \ref{fig: MiDaS figures}, the depth image reports a flat image and it is difficult to understand the depth of the fruit. So, the network cannot effectively estimate the position of the fruit and reports errors up to \SI{10}{\centi\metre} (Fig. \ref{fig: MiDaS residuals}). However, depth essays and calibration procedures should be made to purposefully conclude the non-effectiveness of MiDaS to estimate the depth of the image, i.e., MiDaS requires a complete RGB-D system to correctly calibrate the RGB sensor and network. Despite this error, deep learning-based solutions are much more computing demanding and less straightforward, difficulting to improve the results and track the origin of the errors. Besides, they also require much training and reliable data. On the other hand, MonoVisual3DFilter is data-independent, and its behaviour is more predictable.

Such as identified during the literature review in the introduction section are algorithms such as the SilhoNet, Nerf-Pose, or the GDR-Net. All of these algorithms have detection errors of about \SI{2}{\centi\metre}. Near the MonoVisual3DFilter is the Imitrob model that can estimate the pose of the objects without their model, but reaches an estimation error of \SI{6.5}{\centi\metre} on average. These solutions are competitive with the MonoVisual3DFilter, mainly if we consider complex datasets such as the one where they were essayed. Besides these algorithms can estimate the 6D pose of the target objects. Although these advantages are against the MonoVisual3DFilter, almost all of the solutions are model-dependent and computing-demanding. The MonoVisual3DFilter that we are here proposing is model-independent and only requires a mechanism capable of accurately detecting the target objects in a 2D scene. Besides, from these approaches, we can also conclude that our solution can be improved if we provide it with instance segmentation masks instead of rectangular bounding boxes, that contain areas that do not belong to the objects.  

\begin{figure}[!b]
    \centering
    \includegraphics[width=0.5\textwidth]{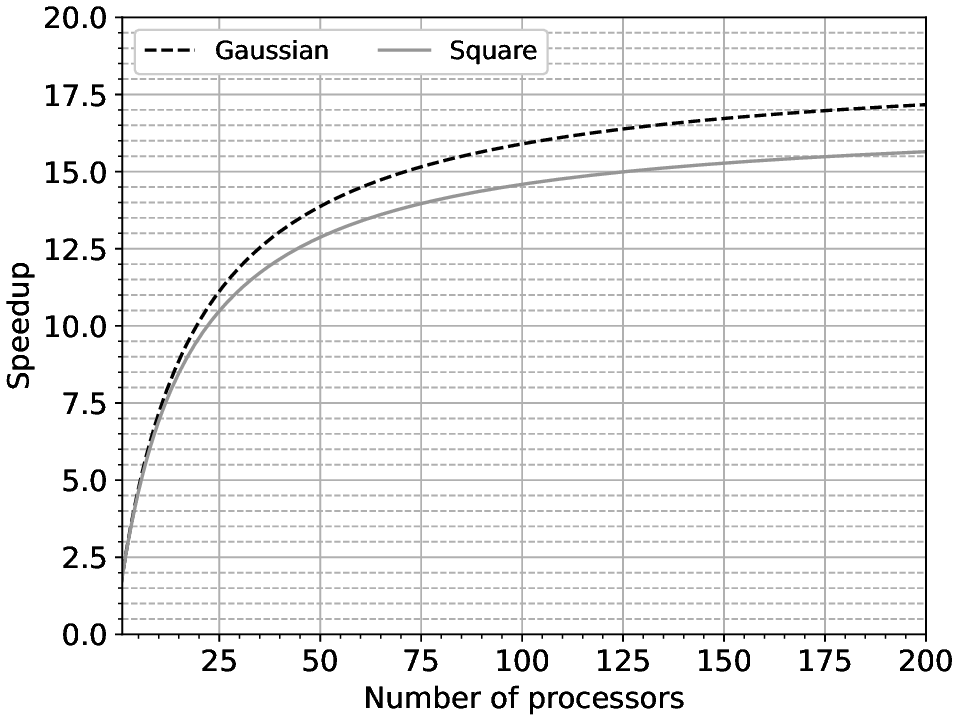}
    \caption{Maximum speedup analysis for parallelisation according to Amdahl's Law for the Gaussian and Square kernels.}
    \label{fig:speedup-plot}
\end{figure}

The MonoVisual3DFilter can estimate objects' position in the \ac{3D}, mainly the circular ones. However, this algorithm is, currently, computationally demanding and requires many resources and time to compute a solution. In our case, the system took about one minute in each pose to compute the decomposed state-space using an Intel Core i7 with \SI{8}{GB} of \ac{ram}. However, this is a highly parallelisable algorithm once all the positions are independent and can be computed simultaneously. So parallel implementation of the MonoVisual3DFilter at the \ac{CPU}, \ac{GPU}, or mainly at the \ac{fpga} can proportionally boost the speed of inference. {Additionally, better optimisation of the implemented code can be done, mainly by using more efficient programming languages such as C or C++ instead of Python, which is less efficient and is interpreted during execution.}

To better understand the advantages of parallelising the MonoVisual3DFilter, we studied the algorithm speedup through Amdahl's law \cite{Amdahl1967} \eqref{eq:amdahl_law}. In this equation, \eqref{eq:amdahl_law}, $\sigma(n)$ is the inherently sequential computations, $\varphi(n)$ is the potentially parallel computations, and $p$ is the number of processors (processes computing in parallel). It is important to mind that Amdahl's law ignores the communications between processes, so this law only computes the maximum speedup, $\Psi(n,p)$. Fig. \ref{fig:speedup-plot} illustrates the maximum reachable speedup by parallelising the histogram filter. The Gaussian filter reaches a higher speedup of about \num{17.5}, but the square is a simpler kernel that operates faster, so the speedup is limited by the inherently sequential operations that cannot be optimised. During simulation, in a computer with an Intel Core i7 and \SI{8}{GB} of \ac{ram}, the histogram filter took about \SI{115}{\second\per{pose}} using the Gaussian Filter and \SI{99}{\second\per{pose}} using the square kernel{, without parallelisation}. The number of available cores also limits the speedup of the histogram filter. Once it is a very parallelisable algorithm, it benefits from using many cores, so the \ac{CPU} is not so interesting such as \ac{GPU} or \ac{fpga}, because it is usually limited to 16 cores. 

\begin{equation}
    \Psi(n,p) \leqslant \frac{\sigma(n)+\varphi(n)}{\sigma(n)+\dfrac{\varphi(n)}{p}}
    \label{eq:amdahl_law}
\end{equation}

\section{Conclusion}
\label{sec: conclusion}

During this experiment, we designed a histogram filter-based algorithm, the MonoVisual3DFilter, to infer the \ac{3d} position of tomatoes in the tomato plant canopy using monocular cameras. The algorithm performed reasonably with an overall error of about \SI{20}{\milli\metre} in laboratory-controlled conditions.

Despite the MonoVisual3DFilter being valid for estimating the tomatoes' position, some additional improvements are required. The next steps should focus on optimising the selection of the observation poses, making these poses adjustable and variable according to the fruit being analysed, and maximising observability through intelligent algorithms. The proposed algorithm can probably be improved if we feed it with instance segmentation masks instead of bounding boxes. Besides, improvements in execution time are still needed by optimising the developed code and implementing parallelisation strategies.

Other opportunities can also be explored while using the proposed algorithm. An example of that is using radar technology that allows one to perceive occluded objects behind the scene. Other similarly perceiving sensors can also be considered, acquiring the system with more robust sensors to perturbances in the scene.

Therefore, using histogram filters to estimate the position of objects, namely the MonoVisual3DFilter, is viable and suitable for operating in the field under controlled scenarios. Further essays should be conducted on real scenarios and the implementation of active perception strategies. 

\bibliographystyle{roblike}
\bibliography{refs}

\begin{con}
\ctitle{Author Contributions}
Conceptualization, S.C.M. and F.N.d.S.; funding acquisition, S.A.M and F.N.d.S.; investigation, S.C.M.; methodology, S.C.M.; software, S.C.M; formal analysis, S.C.M.; resources, F.N.d.S and A.P.M.; project administration, S.A.M, F.N.d.S., J.D. and A.P.M; supervision, F.N.d.S., J.D. and A.P.M.; validation, F.N.d.S., J.D. and A.P.M.; writing---original draft, S.C.M.; writing---review and editing, S.C.M., F.N.d.S., J.D. and A.P.M.

\ctitle{Financial Support}
This work is co-financed by Component 5 -- Capitalization and Business Innovation, integrated in the Resilience Dimension of the Recovery and Resilience Plan within the scope of the Recovery and Resilience Mechanism (MRR) of the European Union (EU), framed in the Next Generation EU, for the period 2021--2026, within project PhenoBot-LA8.3, with reference PRR-C05-i03-I-000134-LA8.3.

Sandro Costa Magalhães is granted by the Portuguese Foundation for Science and Technology (FCT---Fundação para a Ciência e Tecnologia), through the European Social Fund (ESF) integrated into the Program NORTE2020, under scholarship agreement SFRH/BD/147117/2019 (\href{http://dx.doi.org/10.54499/SFRH/BD/147117/2019}{DOI:10.54499/SFRH/BD/147117/2019}).

\ctitle{Conflicts of Interest}
The authors declare no conflicts of interest exist.

\ctitle{Ethical Approval}
Not applicable.
\end{con}

\end{document}